\documentclass[preprint,sort,compress,12pt]{elsarticle}

\usepackage{graphicx}
\usepackage{amsmath}
\usepackage{amssymb}
\usepackage{mathrsfs}
\usepackage{array}
\usepackage{upgreek}
\usepackage{float}
\usepackage{adjustbox}
\usepackage[dvipsnames]{xcolor}
\usepackage[margin=1in]{geometry}
\usepackage[small,compact]{titlesec}
\usepackage[normal,bf,labelsep=colon,skip=2pt]{caption}
\usepackage[titles]{tocloft}
\usepackage{bold-extra}
\usepackage[breaklinks=true]{hyperref}
\usepackage[ruled,vlined]{algorithm2e}
\usepackage{caption}
\usepackage{multirow}
\usepackage{lineno}
\usepackage{tabularx}
\usepackage{diagbox}
\usepackage{algorithmic}

\newcommand{\Tau}{\mathrm{T}}

\DeclareMathOperator*{\argmin}{arg\,min}

\usepackage{xcolor}
\hypersetup{
    colorlinks,
    linkcolor={red!50!black},
    citecolor={blue!50!black},
    urlcolor={blue!80!black}
}

\begin{document}
\begin{frontmatter}
\title{Data-Driven Stochastic Closure Modeling via Conditional Diffusion Model and Neural Operator}

\author[]{Xinghao Dong}
\author[]{Chuanqi Chen}
\author[]{Jin-Long Wu\corref{cor1}} \ead{jinlong.wu@wisc.edu} 
\cortext[cor1]{Corresponding author}

\address{Department of Mechanical Engineering, University of Wisconsin–Madison, Madison, WI 53706}

\begin{abstract}
Closure models are widely used in simulating complex multiscale dynamical systems such as turbulence and the earth system, for which direct numerical simulation that resolves all scales is often too expensive. For those systems without a clear scale separation, deterministic and local closure models usually lack enough generalization capability, which limits their performance in many real-world applications. In this work, we propose a data-driven modeling framework for constructing stochastic and non-local closure models via conditional diffusion model and neural operator. Specifically, the Fourier neural operator is incorporated into a score-based diffusion model, which serves as a data-driven stochastic closure model for complex dynamical systems governed by partial differential equations (PDEs). We also demonstrate how accelerated sampling methods can improve the efficiency of the data-driven stochastic closure model. The results show that the proposed methodology provides a systematic approach via generative machine learning techniques to construct data-driven stochastic closure models for multiscale dynamical systems with continuous spatiotemporal fields.
\end{abstract}


\end{frontmatter}

\section{Introduction}
Complex dynamical systems are ubiquitous in a wide range of scientific and engineering applications, e.g., climate change prediction, renewable energy harvesting, and the discovery of materials for clean fuels and energy storage. These complex systems often feature a wide range of temporal and spatial scales, for which accurately resolving and coupling all of them is still infeasible for many real-world applications~\cite{moin1998direct}. In practice, classical closure modeling approaches empirically avoid resolving some of the scales for affordable numerical simulations by introducing a closure model that accounts for the unresolved scales based on the resolved ones, such as Reynolds Averaged Navier-Stokes (RANS) \cite{launder1983numerical, wilcox1998turbulence} and large eddy simulation (LES) closure models \cite{smagorinsky1963general, deardorff1970numerical}. However, a fundamental challenge in modeling the impacts of those small scales on large scales still exists when the complex system lacks a clear separation between resolved scales and unresolved ones. In such cases, the unresolved (or subgrid) dynamics often interact with the resolved scales in a highly nonlinear and non-local manner, leading to memory effects and feedback loops that are not easily captured by conventional deterministic closures. For instance, in multiscale turbulent flows without a distinct scale gap, the resulting closure terms must incorporate non-Markovian influences—where past states affect the current dynamics—in addition to spatial non-locality~\cite{sanderse2024scientific}. This challenge reveals the limitations of classical closure models that often assume a deterministic form based solely on local information, thereby neglecting crucial non-local interactions; it motivates the development of stochastic models that can effectively incorporate both local and non-local effects~\cite{zwanzig2001nonequilibrium, Lucarini14a,chorin2015discrete,franzke2015stochastic,lu2017data,brennan2018data,palmer2019stochastic,callaham2021nonlinear,schneider2021learning,chen2023stochastic,chen2023simple,mou2023efficient,yang2025active,wu2024learning}.

More recently, researchers have investigated the use of machine learning (ML) to enhance—or even replace—traditional closure models that rely on limited domain knowledge and empirical calibration. By harnessing the rapidly growing volume of available data, these ML methods enable the development of more sophisticated, data-informed closures. A range of approaches has been explored, including sparse dictionary learning~\cite{brunton2016discovering, lu2022discovering, shi2021acd, chen2023ceboosting}, Gaussian process regression~\cite{wan2017reduced, lee2020coarse, papaioannou2022time}, diffusion maps~\cite{galaris2022numerical, liu2015equation, koronaki2020data}, and deep learning techniques~\cite{maulik2019subgrid, gupta2021neural, gupta2023generalized, chen2023operator, chen2024cgnsde, chen2024cgkn, agrawal2024probabilistic, fabiani2024task}. In order to mitigate the “black box” nature of many ML models and improve their interpretability and generalizability, recent studies have emphasized the integration of known physical laws into the closure modeling process. For example, foundational physics underlying turbulent flows were analyzed in~\cite{girimaji2024turbulence}, and the importance of targeted, physics-informed strategies was stressed for capturing the nuances of turbulence. At the same time, efforts continue to be made to incorporate physical constraints during model design and training~\cite{kashinath2021physics, wang2017physics, wu2018physics}. Comprehensive reviews of machine learning-based closure models can be found in~\cite{duraisamy2019turbulence, melchers2023comparison}. 
Despite these advances, most existing works still assume a deterministic model form, which may not adequately capture closure behavior when the resolved and unresolved scales do not exhibit a clear separation~\cite{palmer2019stochastic}.

On the other hand, generative models have distinguished themselves as powerful tools in machine learning tasks such as generating texts, images, and videos. Designed to approximate the underlying distribution of the training dataset and to sample from it efficiently, the generative models can also be used in the stochastic modeling of complex dynamical systems~\cite{stinis2019enforcing,wu2020enforcing,yang2020physics,cheng2020data,perezhogin2023generative}. One standout subclass within generative models is diffusion models~\cite{ho2020denoising, sohl2015deep, dhariwal2021diffusion, song2019generative, song2020score}, which have demonstrated exceptional promise beyond traditional applications such as image and speech synthesis. The general concept of diffusion models is to gradually add noise to the target distribution via a diffusion process, which has a corresponding reverse Markov process that transforms the random noise back into a sample in the target distribution. The foundations of diffusion models can be traced to score matching~\cite{hyvarinen2005estimation} and its connection to denoising autoencoders~\cite{vincent2011connection}, which laid the groundwork for various types of diffusion models. Key developments in diffusion models include denoising diffusion probabilistic models (DDPM)~\cite{ho2020denoising}, which formalized the training and sampling processes, and denoising diffusion implicit models (DDIM)~\cite{song2021denoising}, which introduced faster deterministic sampling. In addition, score-based generative models were proposed by~\cite{song2019generative}, and their unification with diffusion models through score-based SDEs~\cite{song2020score, huang2021variational} has further advanced the field.

Compared to other generative models like Variational Autoencoders (VAEs) \cite{kingma2013auto} and Generative Adversarial Networks (GANs) \cite{goodfellow2014generative}, diffusion models are not constrained by the need for specific model architectures to maintain a tractable normalizing constant, nor do they rely on the often unstable adversarial training methods. The structural flexibility of diffusion models allows for the incorporation of input conditions that can guide the generation process~\cite{saharia2022palette, rombach2022high, ho2021classifierfree}. All these features of diffusion models highlight their potential for advancing the modeling of complex dynamical systems. The conditional score-based diffusion models for imputation (CSDI) \cite{tashiro2021csdi} present an innovative method for time series imputation by leveraging score-based diffusion models. For prediction tasks, TimeGrad~\cite{rasul2021autoregressive} leverages diffusion models and serves as an autoregressive model for time series forecasting. Additionally, DiffSTG~\cite{wen2023diffstg} generalizes DDPMs to spatiotemporal graphs. In parallel, significant advances have been achieved in fluid dynamics, where diffusion models are applied both for the direct generation of spatiotemporal fields \cite{yang2023denoising, li2024synthetic, lippe2024pde, liu2024confild} and for reconstructing high-fidelity fields from low-fidelity approximations \cite{shu2023physics, gao2024bayesian, gao2024generative}. For instance, the G-LED \cite{gao2024generative} combines Bayesian diffusion models with multi-head autoregressive attention to accelerate simulations by learning effective dynamics, while CoNFiLD-inlet~\cite{liu2024confild} leverages generative latent diffusion models with conditional neural field encoding to synthesize realistic, stochastic turbulence inflow conditions that capture both coherent structures and multiscale stochasticity. Additionally, the integration of physics constraints and domain knowledge has been investigated in~\cite{qiu2024pi, bastek2024physics, jacobsen2023cocogen}. In particular, CoCoGen~\cite{jacobsen2023cocogen} efficiently enforces physical laws in score-based generative models ensuring generated samples maintain physical consistency by incorporating discretized PDE information during the sample generation stage rather than during training.

However, the use of diffusion models for closure modeling is still prevented by a common drawback of standard diffusion models, i.e., the slow sampling speed, which is mainly because diffusion models often require hundreds or even thousands of inference steps for the generation of each sample. For a closure model, it needs to be evaluated at every numerical time step during the simulation, and thus standard diffusion models are too costly to serve as closure models. To address this drawback of diffusion models, some recent developments in acceleration techniques, such as distillation~\cite{salimans2022progressive, song2023consistency, meng2023distillation} and adaptive numerical schemes~\cite{jolicoeur2021gotta, karras2022elucidating}, can be promising. One of the major goals of this work is to demonstrate the capability of diffusion models as stochastic closure models, utilizing these recent techniques to accelerate the sampling of diffusion models. 

Furthermore, many interesting application areas of closure modeling involve continuous spatiotemporal fields, such as modeling turbulence in fluid dynamics. For these applications, standard diffusion models designed for images with finite resolutions (i.e., in a discretized vector space) can be insufficient. The recent development of continuous extensions of traditional neural networks (NN), known as neural operators~\cite{Lu2021, li2021fourier, kovachki2023neural}, seeks to provide a continuous framework to capture the intrinsic multiscale nature of complex dynamical system states and associated data sources. Unlike traditional methods that are confined to a fixed mesh resolution, the adaptability of neural operators to different resolutions in training and testing enables them to handle data with different temporal and spatial resolutions efficiently. An example of neural operator is deep operator network (DeepONet)~\cite{Lu2021}, which provides an architecture that consists of multiple deep neural networks -- branch nets for encoding input functions and trunk nets for the output. On the other hand, the Fourier neural operator (FNO)~\cite{li2021fourier} leverages Fourier transforms to construct a framework of approximating nonlinear operators. The merits of operator learning have been demonstrated in various applications involving partial differential equations such as solid mechanics, fluid dynamics, weather forecasting, and other fields where continuous spatiotemporal fields are of interest and capturing long-range interactions is essential~\cite{chen2023operator,tran2023factorized,goswami2022physics,wen2022u,li2023fourier}.

In this paper, we introduce a novel stochastic closure modeling framework for complex dynamical systems governed by PDEs. This modeling framework builds on the strengths of conditional diffusion models and operator learning techniques. More specifically, a score function constructed by the Fourier neural operator is trained to approximate the unknown score function that corresponds to a conditional probability distribution of the closure term. The score-based conditional diffusion model then serves as the data-driven stochastic closure and is deployed in the numerical simulations of the classical physics-based solver as an additional correction term. The key highlights of this work are summarized below:

\begin{itemize}
\item We develop a data-driven stochastic closure modeling framework for complex dynamical systems governed by PDEs by training a score-based generative model that captures the conditional probability distribution of the unknown closure terms with respect to the known information, e.g., numerically resolved system states, sparse measurements of the true closure terms, and estimations from an existing physics-based closure model.

\item We explore the Fourier neural operator to construct the score function in the generative model, which leads to two advantages: (i) the resolution invariance property that facilitates using data with various resolutions, and (ii) the non-local property that goes beyond the assumption of a local closure model.

\item We proposed a rapid sampling strategy that accelerates the sampling speed of the data-driven stochastic closure model by up to a factor of $O(100)$, significantly enhancing the practicality of diffusion models being used in the context of stochastic closure modeling.
\end{itemize}

\section{Methodology}
The governing equation of a dynamical system in general form studied in this work can be written as:
\begin{equation}
    \label{eqn:true_system}
    \frac{\partial v}{\partial t} = M(v),
\end{equation}
where $v$ denotes the system state and $M$ represents the dynamics derived from first principles, which can involve highly nonlinear mappings, differential operators, and integral operators. In practice, numerically resolving every detail of $v$ can be infeasible, which motivates the focus on a reduced-order system:
\begin{equation}
    \label{eqn:model_system}
    \frac{\partial V}{\partial t} = \widetilde{M}(V),
\end{equation}
where $V:=\mathcal{K}(v)$ denotes a reduced-order representation of $v$ with a map $\mathcal{K}$ that extracts the information to be numerically resolved, and $\widetilde{M}(V)$ is derived based on domain knowledge to approximate the reduced-order dynamics $\mathcal{K} \circ M(v)$ so that a closed system of $V$ is obtained for practical numerical simulations. To enhance the performance of the classical closure models $\widetilde{M}$, researchers have recently explored introducing a machine-learning-based correction term $U$, with the aim that $\widetilde{M}(V)+U$ can potentially better characterize the actual reduced-order dynamics. 

In this work, we focus on a stochastic form of the correction term $U$ enabled by the recent developments of diffusion models. More specifically, we aim to approximate the distribution $p(U)$ and to efficiently sample from it. It should be noted that $U$ often depends on the current state $V$ and potentially other information, such as the temporal history of $V$ or the spatially non-local information. Therefore, we denote all the dependent information of a well-characterized correction term $U$ as $\mathrm{y}$ and focus on a conditional diffusion model framework that targets the approximation and efficient sampling of $p(U \mid \mathrm{y})$. 

Effectively, the conditional diffusion model framework aims to describe the dynamics of $U$ via a stochastic partial differential equation (SPDE):
\begin{equation}
\frac{\partial U}{\partial t} = h(U;\ \mathrm{y}) + \xi,
\end{equation}
where $\xi$ denotes space-time noise, and $h$ often involves partial differential operators, with the assumption that the regularity requirements are satisfied so that the conditional probability distribution $p(U \mid \mathrm{y})$ is well defined at each time $t$. Instead of directly modeling the unknown operator $h$ and the space-time noise $\xi$ for numerically simulating the above SPDE, the proposed framework aims to characterize and sample the distribution $p(U \mid \mathrm{y})$ via a conditional diffusion model.

\begin{figure}[H]
    \centering
    \includegraphics[width = \linewidth]{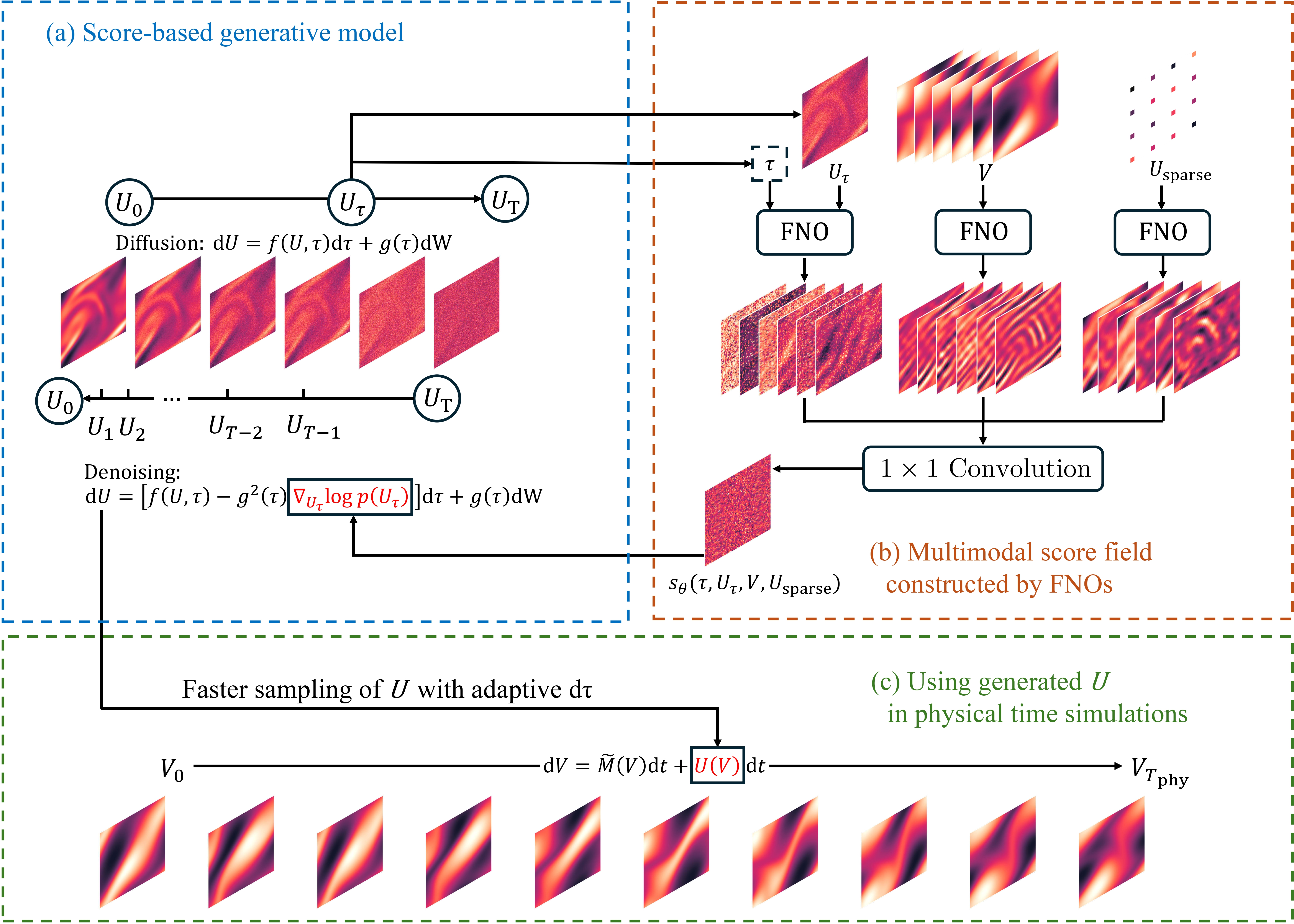}
    \caption{A schematic diagram of the proposed framework for stochastic closure modeling via score-based generative model and Fourier neural operator. The score function with multimodal inputs is constructed by FNOs and deployed in a conditional diffusion model, with which samples can be generated to serve as a data-driven closure term for a complex dynamical system whose dynamics are only partially known or even completely unknown.}
    \label{fig:schematic}
\end{figure}

The proposed stochastic closure modeling framework is illustrated in Fig.~\ref{fig:schematic}, with key components further introduced in Sections~\ref{ssec:method_SBGM} to \ref{ssec:method_fast_sampling}. The associated algorithms are described in~\ref{sec: algorithms}.

\subsection{Score-Based Generative Model}
\label{ssec:method_SBGM}
Instead of directly approximating the probability distribution function, score-based models leverage available independent identically distributed (i.i.d.) data samples to learn the score function $\nabla_U \log p(U)$, defined as the gradient of the log probability density function~\cite{chwialkowski2016kernel, liu2016kernelized}. Unlike the probability density function, the score function does not require a tractable normalizing constant and can be trained with samples of $U$ by score matching~\cite{vincent2011connection, song2020sliced}. Once the score function is trained, the sampling from $p(U)$ can be achieved via Langevin dynamics~\cite{parisi1981correlation, grenander1994representations}, usually beginning with samples from an easy-to-sample distribution, e.g., a Gaussian distribution. In practice, $U$ is often high-dimensional, and it is infeasible to have enough data to guarantee an accurate trained score function within the entire high-dimensional space.

To address the challenge of inaccurate score estimation in some regions of low-density or sparse data~\cite{song2019generative}, one strategy involves progressively adding noise to the initial distribution and training the score-based model on these perturbed data points. In a continuous-time framework, the process of noise adding is achieved by a stochastic differential equation (SDE)~\cite{song2020score}:
\begin{align}\label{eqn: forwardSDE}
\mathrm{d} U = f(U, \tau) \mathrm{d}\tau + g(\tau)\mathrm{dW}.
\end{align}
Here, $\mathrm{W}$ denotes a Wiener process, and $\tau \in [0, \Tau]$ represents the time in the SDE simulation, which is usually incorporated as an input into the score-based model as an indicator of noise levels. The choice of the detailed forms for $f$ and $g$ in this SDE is not unique. A common choice is known as the variance exploding (VE) SDE \cite{song2020score}:
\begin{align}\label{eqn: VESDE}
    \mathrm{d}U = \sigma^\tau \mathrm{dW},
\end{align}
with $\sigma$ as a constant scalar. This is equivalent to defining a Gaussian transition kernel in this form:
\begin{align}\label{eqn: transitionkernel}
    p(U_\tau \mid U_0) = \mathcal{N}\left(\mu(U, \tau), \Sigma(\tau)\right),
\end{align}
where
\begin{align}\label{eqn: transitionkernelparam}
    \mu(U, \tau) &= U_0, \\
    \Sigma(\tau) &= \frac{1}{2 \log \sigma}\left(\sigma^{2\tau}-1\right) I.
\end{align}

At $\tau = 0$, the distribution $p(U_0)$ corresponds to the true data distribution. With the transition kernel defined in Eq.~\eqref{eqn: transitionkernel}, the final distribution after the diffusion process introduced in Eq.~\eqref{eqn: VESDE} can also be derived analytically:
\begin{align}\label{eqn: prior}
    p(U_\Tau) = \int p(U_0) \cdot \mathcal{N}\left(U_0, \frac{1}{2 \log \sigma}\left(\sigma^{2\Tau}-1\right) I\right) \mathrm{d}U_0 \approx \mathcal{N}\left(0, \frac{1}{2 \log \sigma}\left(\sigma^{2\Tau}-1\right) I\right).
\end{align}
When $\tau = \Tau$ and $\sigma$ are sufficiently large to ensure enough amount of noise is added, $p(U_\Tau)$ approaches a zero-mean Gaussian distribution, which is easy to sample from. With samples from the distribution $p(U_\Tau)$, the samples from the target distribution $p(U_0)$ can be obtained by solving the following reverse SDE~\cite{anderson1982reverse}:
\begin{align}\label{eqn: reverseSDE}
\mathrm{d} U = \left[f(U, \tau) - g^2(\tau)\nabla_{U_\tau} \log p(U_\tau)\right] \mathrm{d}\tau + g(\tau)\mathrm{dW},
\end{align}
from $\tau=T$ to $\tau=0$, and the score function $\nabla_{U_\tau} \log p(U_\tau)$ can be approximated by a neural network, which can be trained via score matching. The standard score matching and the denoising score matching~\cite{vincent2011connection, song2020sliced} have the following equivalency:
\begin{align}\label{eqn: ESMDSM}
    \mathbb{E}_{U_\tau \sim p(U_\tau)} \left \| \nabla_{U_\tau} \log p(U_\tau) - s_\theta \right \|^2_2 = \mathbb{E}_{U_\tau \sim p(U_\tau \mid U_0)}\mathbb{E}_{U_0 \sim p(U_0)} \left \| \nabla_{U_\tau} \log p(U_\tau \mid U_0) - s_\theta\right \|^2_2 + C,
\end{align}
where the left-hand and right-hand sides correspond to the explicit and denoising score matching, respectively, with $s_\theta$ representing the model to be trained and $C$ being a constant independent of $\theta$. It is worth noting that the true score function required by the explicit score matching is often inaccessible, and thus denoising score matching provides a practical way of obtaining a neural network $s_\theta$ that approximates the true score function. With an adequately expressive model and a sufficient enough dataset, the above equivalency ensures that a score-based model $s_\theta(\tau, U_\tau)$ trained with the following denoising score matching objective function
\begin{align}\label{eqn: unconscorematching}
    \theta^* = \argmin_\theta \left \{ \mathbb{E}_{\tau \sim \mathcal{U}(0, \Tau)} \mathbb{E}_{U_\tau \sim p(U_\tau \mid U_0)} \mathbb{E}_{U_0 \sim p(U_0)} \left \| \nabla_{U_\tau} \log p(U_\tau \mid U_0) - s_\theta(\tau, U_\tau)\right \|^2_2\right \}
\end{align}
will effectively approximate the score function $\nabla_{U_\tau} \log p(U_\tau)$ for almost all $U_\tau$ and $\tau\in [0,\Tau]$. Detailed proofs can be found in~\ref{ssec: proof_ESMDSM}.

In this work, we focus on the use of score-based diffusion models for closure modeling, which aims to provide a correction term $U$ to the right-hand side of Eq.~\eqref{eqn:model_system}, such that the numerical simulation would lead to resolved system state $V$ that has a better agreement with the true one. In many applications, it is expected that the correction term $U$ should depend on the resolved system state $V$ and can benefit from the knowledge of some other information, such as the estimation of $U$ from existing physics-based models or sparse measurements of true $U$ from the real system. Therefore, a conditional diffusion model is needed to account for all the dependent information to better characterize the correction term $U$ for the modeled system in Eq.~\eqref{eqn:model_system}. The conditional score-based model~\cite{gao2024bayesian, tashiro2021csdi, shu2023physics, bastek2024physics} is designed to learn the score function of a conditional distribution $p(U \mid \mathrm{y})$, which is represented as $\nabla_{U} \log p(U \mid \mathrm{y})$. Here, $\mathrm{y}$ denotes the dependent information in general that is helpful for characterizing the closure model. 

Challenges in transitioning to the conditional model involve expectations of the conditional transitional kernel $U_\tau \sim p(U_\tau \mid U_0, \mathrm{y})$ and a conditional initial distribution $U_0 \sim p(U_0\mid \mathrm{y})$. The forward SDE in a diffusion process operates as a Markov chain, meaning the transition kernel is unaffected by any input conditions $\mathrm{y}$, which implies that $p(U_\tau \mid U_0, \mathrm{y}) = p(U_\tau \mid U_0)$. In the objective function, we choose to take the expectation of a more tractable joint distribution $p(U_0, \mathrm{y})$, instead of the conditional distribution  $p(U_0\mid \mathrm{y})$. Detailed derivations are provided in~\ref{ssec: CSM}, and the objective function in the conditional model setting is presented below:
\begin{align}\label{eqn: conditionscorematching}
    \theta^* = \argmin_\theta \left \{ \mathbb{E}_{\tau \sim \mathcal{U}(0, \Tau)} \mathbb{E}_{U_\tau \sim p(U_\tau \mid U_0)} \mathbb{E}_{U_0, \mathrm{y} \sim p(U_0, \mathrm{y})} \left \| \nabla_{U_\tau} \log p(U_\tau \mid U_0) - s_\theta(\tau, U_\tau, \mathrm{y})\right \|^2_2\right \}.
\end{align}

Once the score function $s_\theta(\tau, U_\tau, \mathrm{y})$ is trained based on the above objective function, the sampling process involves solving the backward SDE defined in Eq.~\eqref{eqn: reverseSDE} from $\tau = \Tau$ to $\tau = 0$, with the score function being replaced by the conditional one $\nabla_{U_\tau} \log p(U_\tau \mid \mathrm{y})$. Associated with the VE SDE in Eq.~\eqref{eqn: VESDE}, the backward SDE has the analytical form of 
\begin{align}\label{eqn: conditional_reverseSDE}
    \mathrm{d}U = -\sigma^{2\tau} \nabla_{U_\tau} \log p(U_\tau \mid \mathrm{y}) d\tau + \sigma^\tau \mathrm{d}\mathrm{W}.
\end{align}
One common choice for solving Eq.~\eqref{eqn: conditional_reverseSDE} is the Euler-Maruyama numerical scheme, which can be expressed as
\begin{align}\label{eqn: eulermaruyama}
    \begin{split}
        U_{\tau - \Delta \tau} &= U_\tau + \sigma^{2\tau} \nabla_{U_\tau} \log p(U_\tau \mid \mathrm{y}) \Delta \tau + \sigma^\tau \sqrt{\Delta \tau} z \\
        &\approx U_\tau + \sigma^{2\tau} s_\theta(\tau, U_\tau, \mathrm{y}) \Delta \tau + \sigma^\tau \sqrt{\Delta \tau} z,
    \end{split}
\end{align}
with $z \sim \mathcal{N}(0, I)$, where the true score function is approximated by the trained neural network model $s_\theta$, which depends on the noise level $\tau$, the intermediate noisy state $U_\tau$, and other conditional inputs $\mathrm{y}$. 

\subsection{Neural Operator for the Construction of Score Function}\label{ssec:operator_learning}
In this work, we focus on the complex dynamical systems governed by PDEs, whose true system state $v$ in Eq.~\eqref{eqn:true_system} and the resolved system state $V$ in Eq.~\eqref{eqn:model_system} are both continuous spatiotemporal fields. Instead of using a standard diffusion model designed for images in finite vector spaces, a Fourier neural operator~\cite{li2021fourier} is employed to construct the score function, which not only enables a generative diffusion model for continuous spatiotemporal fields but also facilitates the learning of a non-local closure model if needed. It is worth noting that non-locality in space and/or time is important for the modeling of many complex dynamical systems and the general concept of non-local models can be found in~\cite{du2012analysis,d2020numerical,you2021data,ma2019model,wang2020recurrent,lin2021data,charalampopoulos2022machine,wu2024learning}.

The standard Fourier neural operator is represented by a sequence of $N$ Fourier layers, with the detailed form of the $n$-th layer defined below:
\begin{align}\label{eqn: FT}
    Q_{n+1}(\mathrm{x}) = \sigma\left(LQ_n\left(\mathrm{x}\right) + \left(KQ_n\right)\left(\mathrm{x}\right)\right),  \quad n = 0, 1, \cdots, N-1,
\end{align}
where $\sigma$ is the activation function and $L$ is a linear transformation. $K$ is defined as the Fourier integral operator so that $(KQ_n)(\mathrm{x}) = \mathcal{F}^{-1}\left(R \cdot (\mathcal{F}Q_n)\right)(\mathrm{x})$ with $R$ being the complex-valued parameters in the neural network, and $\mathcal{F}$ with $\mathcal{F}^{-1}$ denoting the Fourier transform and inverse Fourier transform, respectively. 

As shown in Fig.~\ref{fig:schematic}, FNOs are used to construct the score-based model $s_\theta(\tau, U_\tau, \mathrm{y})$, which takes noise level $\tau$, corresponding perturbed target field $U_\tau$, and a set of conditions $\mathrm{y}$ as inputs. Our architecture involves a combination of multiple FNO pipelines, which are designed to handle different inputs to the score-based model. 

The noise levels, indicated by the SDE time information $\tau$, are incorporated via Gaussian random features as described in \cite{tancik2020fourier}. For a given SDE time scalar $\tau$, it is processed by Gaussian Fourier projection in the form of
\begin{equation}\label{eqn: GRF}
 r(\tau) = \left[\sin(2\pi \gamma \tau); \cos(2\pi \gamma \tau)\right],
\end{equation}
where $\left[\cdot; \cdot \right]$ denotes the concatenation of two vectors, and $\gamma$ represents fixed, non-trainable Gaussian weights that are randomly sampled during initialization. The features $r(\tau)$ are then further processed by applying an activation function, followed by a linear layer and a dense layer, resulting in a rich embedding that can be incorporated into the FNO pipelines. This allows the model to learn the dependency between noise levels $\tau$ and corresponding diffused states $U_\tau$. More details on the construction of the FNOs, including the choice of activation functions and hyper-parameters can be found in \ref{sec: train_detail}.

FNO is suitable to construct resolution-invariant and non-local models that involve spatially continuous fields, and we construct an FNO-based multimodal score function with three types of inputs: (i) the associated intermediate diffusion states $U_\tau$, (ii) the related resolved system states $V$, and (iii) other useful information to characterize the desired closure term $U$, such as the sparse measurements of true $U$ or the estimation of closure term via physics-based models as additional conditional inputs in $\mathrm{y}$. In practice, the sparse information of true closure term $U$ may not be directly measurable, but it can still be estimated based on sparse measurements of the original system state $v$. In terms of the associated intermediate diffusion states $U_\tau$, the resolved system states $V$, and the estimation of closure term via physics-based models, they often share the same spatial resolution as the desired true closure term, and thus standard FNOs can be employed to handle each of them. On the other hand, the sparse measurements of the true closure term likely have a different spatial resolution and thus require numerical upscaling methods before being fed into a standard FNO as input. Finally, all information obtained from the multiple parallel pipelines of FNOs is concatenated and convolved to produce the final output of the score function estimation, which is then optimized through the denoising score matching shown in Eq.~\eqref{eqn: conditionscorematching}, and an illustration of using FNOs to construct the score function is presented in Fig.~\ref{fig:schematic}.

It is worth noting that the desired correction term $U(\mathrm{x}, t_n)$ can also depend on a time series of its historical snapshots 
\begin{equation}
    \left\{ U(\mathrm{x}, t) \mid t \in [t_m, t_n) \text{ with } t_m, t_n \in [0, T_{\text{phy}}] \text{ and } t_m < t_n\right\},
\end{equation}
and a time series of historical data of resolved system states $V$: \begin{equation}
    \left\{ V(\mathrm{x}, t) \mid t \in [t_m, t_n) \text{ with } t_m, t_n \in [0, T_{\text{phy}}] \text{ and } t_m < t_n\right\},
\end{equation}
where $t$ and $T_{\text{phy}}$ are defined as the physical time. To handle time series of spatial fields as conditional inputs, a 3-D FNO \cite{li2021fourier} that convolves both spatially and temporally can be utilized.

\subsection{Fast Sampling Strategies}\label{ssec:method_fast_sampling}
If $\tau=\Tau$ and $\sigma$ are sufficiently large, the prior distribution in Eq.~\eqref{eqn: prior} becomes an easy-to-sample Gaussian distribution. Therefore, the sampling process from the desired distribution by using a score-based diffusion model only requires numerically simulating the reverse SDE in Eq.~\eqref{eqn: conditional_reverseSDE}. However, this simulation-based sampling process can be too costly when it serves as a stochastic closure model, which demands a new sample generated by the conditional diffusion model for each time step in the simulation of the model system in Eq.~\eqref{eqn:model_system}. For instance, the backward SDE using the Euler-Maruyama solver in Eq.~\eqref{eqn: eulermaruyama} often requires thousands or even more time steps to ensure the quality of a generated sample.

In this section, we introduce a fast sampling strategy that speeds up the sampling process by a factor of $O(100)$. This substantial improvement is partly due to the adoption of adaptive time step size in Euler-Maruyama solvers, inspired by the findings of Karras et al.~\cite{karras2022elucidating}. Instead of using uniform step sizes, we utilize a scheduling function to monotonically decrease the step sizes as the simulation progresses from $\tau = \Tau$ to $\tau = 0$. The scheduling function is defined as:
\begin{align}\label{eqn: adaptive_schedule}
    \tau_{i < N} = \left(\tau_{\max}^{\frac{1}{\rho}} + \frac{i}{N-1} \left(\tau_{\min}^{\frac{1}{\rho}} - \tau_{\max}^{\frac{1}{\rho}}\right)\right)^\rho
\end{align}
with $\rho = 7$, $\tau_{\max} = \Tau$, $\tau_{\min} = 10^{-3}$, $\tau_N = 0$, and N is the number of desired steps. In the numerical scheme outlined in Eq.~\eqref{eqn: eulermaruyama}, although larger step sizes can introduce greater numerical errors, the quality of a generated sample at the end is not sensitive to those errors due to the high noise levels at earlier stages. The noise at earlier stages of the backward SDE effectively dominates the errors introduced by larger time steps, as changes in the system state are mainly dictated by the added noise through the forward SDE, rather than the accuracy of the numerical solver for the backward SDE. As $\tau$ decreases and noise levels gradually diminish, the accuracy of the backward SDE solver becomes increasingly crucial, which demands smaller time steps. The adaptive time-stepping approach in Eq.~\eqref{eqn: adaptive_schedule} ensures that the solver's performance balances efficiency and accuracy throughout the whole simulation of the backward SDE from $\tau=\Tau$ to $\tau=0$. 

Another enhancement to speed up the sampling process is using a smaller $\Tau$ enabled by the incorporation of conditional inputs in the diffusion model. Starting from a prior distribution that is closer to the ground truth data distribution (e.g., $\Tau = 0.1$ instead of $\Tau=1$) helps to reduce the computational cost of solving the backward SDE. The improvement was made possible because the incorporated conditions can mitigate the need to explore the entire high-dimensional data space that usually requires highly perturbed distributions, i.e., larger $\Tau$ for the forward SDE in a diffusion model. When the relationships between conditions and targets are deterministic, the conditional distribution becomes degenerate, allowing for precise sample generation from any starting prior distribution. By focusing on a smaller regime of the data space, conditional models improve the efficiency and accuracy of the generation process, eliminating the need to cover broader data spaces as unconditional models have to do.

\section{Numerical Results}\label{sec: numerical_results}

\subsection{Numerical Example Setup}\label{ssec: setup}

In our numerical examples, we consider the following stochastic 2-D Navier-Stokes equation for a viscous, incompressible fluid in vorticity form on the unit torus:
\begin{equation}\label{2dNS}
    \begin{aligned}
        \frac{\partial \omega(\mathrm{x}, t)}{\partial t} = -u(\mathrm{x}, t) \cdot \nabla\omega(\mathrm{x}, t) &+ f(\mathrm{x}) + \nu\nabla^2\omega(\mathrm{x}, t) + \beta\xi, \quad &&(t, \mathrm{x}) \in (0, T_{\text{phy}}] \times (0, L)^2\\
        \nabla \cdot u(\mathrm{x}, t) &= 0, &&(t, \mathrm{x}) \in (0, T_{\text{phy}}] \times (0, L)^2\\
        \omega(\mathrm{x}, 0) &= \omega_0(\mathrm{x}), &&\mathrm{x} \in (0, L)^2
    \end{aligned}
\end{equation}
which is intended as a prototypical model for turbulent flows encountered in geophysical, environmental, and engineering applications, where a clear separation between resolved and unresolved scales is often absent. Here, $u$ denotes the divergence-free velocity field, and $\omega = \nabla \times u$ represents the corresponding vorticity. We set the viscosity $\nu = 10^{-3}$ and impose periodic boundary conditions on a square domain with length $L=1$. The initial vorticity $\omega(0,\mathrm{x}) = \omega_0(\mathrm{x})$ is generated from a 2-D Gaussian random field $\omega_0 \sim \mathcal{N}\Bigl(0,\, 7^{3/2}(-\Delta + 49I)^{-5/2}\Bigr)$, where $\Delta$ is the Laplace operator and $I$ is the identity operator, ensuring a broad spectrum of scales in the initial state. The deterministic forcing is given by $f(\mathrm{x}) = 0.1\Bigl(\sin(2\pi(x+y)) + \cos(2\pi(x+y))\Bigr)$, which acts as a spatially varying driver that, together with the periodic boundaries, promotes the formation of coherent, large-scale structures. Additionally, the system is perturbed by a small-amplitude stochastic forcing $\xi = \mathrm{d}\mathrm{W}/\mathrm{d}t$ (with $\mathrm{W}$ being a Wiener process) rescaled by $\beta = 5\times10^{-5}$, thereby mimicking the effects of unresolved small-scale dynamics. In this regime, low viscosity combined with moderate deterministic and stochastic forcing yields a turbulent flow where nonlinear convection drives chaotic energy transfers and vortex interactions, capturing the complex interplay between local dissipation and non-local energy transport.

We numerically solve Eq.~\eqref{2dNS} using a pseudo-spectral solver described in~\ref{ssec: pseudospectral} and time is advanced by the Crank-Nicolson method described in \ref{ssec: cranknicolson} with a uniform resolution of $256 \times 256$ and a fixed time step $\Delta t = 10^{-3}$. System states and other needed properties are recorded every 10 time steps, i.e. a physical time unit of $10^{-2}$s, for producing the training and test data. We simulate a total of 10 time series each with a physical temporal length of $40$ seconds but only use the snapshots between $30$s and $40$s for training and testing. This is because the initial period of the simulation, from $0$s to $30$s, is used to allow the system to warm up and reach a more stable state, avoiding the influence of initial transient results. For the training data, we use the first 8 time series, while reserving 2 time series for testing. The high-fidelity data is first evenly sub-sampled to $64 \times 64$. For testing, we evaluate our model's performance and confirm the resolution invariance using three different data resolutions: $64 \times 64$, $128 \times 128$, and $256 \times 256$.

\subsection{Partial Knowledge of the Dynamics and Closure Models}\label{ssec: partial_knowledge}
In this work, we consider two scenarios in which the right-hand side of Eq.~\eqref{2dNS} is only partially known. The missing dynamic in each case is characterized by a different dependency on the resolved system state—either linear and local or nonlinear and non-local.

In the first scenario, we focus on the stochastic viscous diffusion term, defined as
\begin{equation}\label{eqn: diffusion}
    G(\mathrm{x}, t) = \nu\nabla^2\omega(\mathrm{x}, t) + 2\beta\xi.
\end{equation}
Here, the Laplacian operator \(\nabla^2\) exhibits two fundamental properties. First, it is linear, meaning that it obeys the principle of superposition when acting on the function \(\omega\). Second, it is local, so that its evaluation at any spatial point \(\mathrm{x}\) depends solely on the behavior of \(\omega\) in an infinitesimally small neighborhood around \(\mathrm{x}\). To account for the absence of this term in our model, we introduce a closure model by training a conditional diffusion model that captures the contribution of the missing stochastic diffusion effects.

In contrast, in the second scenario, we address the stochastic convection term
\begin{equation}\label{eqn: convection}
    H(\mathrm{x}, t) = -u(\mathrm{x}, t) \cdot \nabla\omega(\mathrm{x}, t) + \beta\xi,
\end{equation}
which exhibits both nonlinear and non-local dependency on the system state $\omega$. The nonlinearity arises because the velocity field $u$, which is derived from $\omega$, is multiplied by the gradient $\nabla \omega$, creating a compounded dependency on $\omega$ itself. In fact, the velocity field $u$ is computed from $\omega$ by solving a Poisson equation—a procedure mandated by the incompressibility condition. Specifically, in an incompressible flow, the vorticity-stream function formulation is often used, where the stream function $\psi$ satisfies
\begin{equation}\label{eqn: poisson}
    \nabla^2 \psi = -\omega,
\end{equation}
and the velocity field is then obtained via
\begin{equation}
    u = \nabla^{\perp} \psi,
\end{equation}
with \(\nabla^{\perp} \psi\) denoting the perpendicular gradient of \(\psi\). Since solving the Poisson equation requires integrating information over the entire spatial domain, the convection term \(H\) inherently involves non-local interactions. To approximate the contribution of this nonlinear, non-local convection term, we construct a separate conditional diffusion model with multiple pipelines of FNOs. FNOs approximate operators by projecting functions onto a global Fourier basis, where each Fourier mode inherently aggregates information from the entire domain. This global representation ensures that any local perturbation in the resolved system state $\omega$ influences its spectral coefficients, thereby allowing the model to capture long-range interactions and complex dynamics associated with the advection process.

To further distinguish and validate the difference between the local and non-local relationships, we introduce a spatial correlation metric based on the Pearson correlation coefficient. Specifically, let $U(\mathrm{x}, t)$ denote the missing dynamic -- representing either the viscous diffusion term $G$ or the nonlinear convection term $H$ -- and let $\omega(\mathrm{x}, t)$ be the vorticity field. Both fields are defined on the spatial domain $\Omega = [0, L]^2$, where $L$ is the physical length of the domain. The center of the domain, chosen as the reference point, is denoted by $\mathrm{x}_c = (\frac{L}{2}, \frac{L}{2})$. For each spatial point $\mathrm{x} \in \Omega$, we define the Pearson correlation coefficient between the time series $U(\mathrm{x}_c, t)$ at the reference point and the time series $\omega(\mathrm{x}, t)$ at $\mathrm{x} \in \Omega$ as
\begin{equation}\label{eqn: spatial_correlation}
    C_{U, \omega}(\mathrm{x}) = \frac{\int_0^{T_{\text{phy}}}\left[U(\mathrm{x}_c, t) - \overline{U}(\mathrm{x_c})\right] \left[\omega(\mathrm{x}, t) - \overline{\omega}(\mathrm{x})\right]\mathrm{d}t}{\sqrt{\int_0^{T_{\text{phy}}}\left[U(\mathrm{x}_c, t) - \overline{U}(\mathrm{x_c})\right]\mathrm{d}t \int_0^{T_{\text{phy}}} \left[\omega(\mathrm{x}, t) - \overline{\omega}(\mathrm{x})\right]\mathrm{d}t}}
\end{equation}
where $T_{\text{phy}}$ is the total simulation time, and the time averages are defined by
\begin{equation}\label{eqn: time_avg}
\overline{U}(\mathrm{x}_c) = \frac{1}{T_{\text{phy}}}\int_0^{T_{\text{phy}}}U(\mathrm{x_c}, t) \mathrm{d}t, \quad \overline{\omega}(\mathrm{x}) = \frac{1}{T_{\text{phy}}}\int_0^{T_{\text{phy}}}\omega(\mathrm{x}, t) \mathrm{d}t.
\end{equation}

This spatial correlation map, \( C_{U,\omega}(x) \), quantifies the similarity between the temporal evolution of the missing dynamic \( U \) at the reference point and that of the vorticity field at every other spatial location. When \( U \) corresponds to the viscous diffusion term \( G \), the correlation map typically exhibits a pronounced peak near \( x_c \). This behavior reflects the local nature of the Laplacian operator, which predominantly depends on the values of \( \omega \) in the immediate vicinity of the point where it is computed; the correlation decays sharply away from the center.

\begin{figure}[H]
    \centering
    \includegraphics[width=0.6\linewidth]{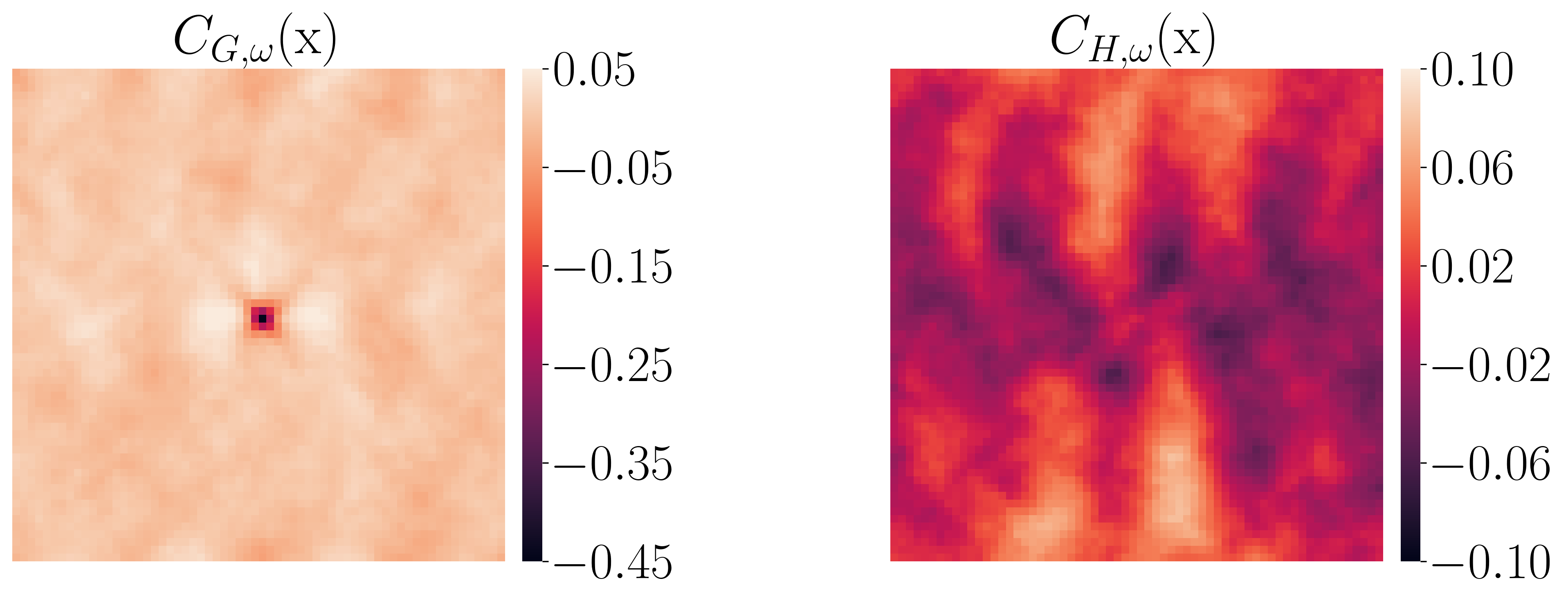} 
    \caption{Left: spatial correlation map $C_{G, \omega}(\mathrm{x})$ calculated between $G(\mathrm{x}_c, t)$ and $\omega(\mathrm{x}, t)$. Right: spatial correlation map $C_{H, \omega}(\mathrm{x})$ calculated between $H(\mathrm{x}_c, t)$ and $\omega(\mathrm{x}, t)$.}
    \label{fig: non-local}
\end{figure}

In contrast, when \( U \) represents the nonlinear convection term \( H \), the correlation map does not show a distinct dominant region. This indicates that the convection process involves a broader range of contributions from the vorticity field, underscoring its inherently non-local character. Figure~\ref{fig: non-local} illustrates these differences in the correlation patterns, providing clear evidence of the local versus non-local nature of the underlying dynamics. For computing the correlation maps below, we first collected a time series of 10,000 snapshots from a single simulation run of Eq.~\eqref{2dNS} up to $T_{\text{phy}}=100$. We then augmented the dataset 15 times by exploiting the periodic boundary conditions, shifting the domain to generate different reference points. Consequently, the final correlation maps represent the averaged correlations computed from a total of 150,000 samples with various reference points.

Throughout both cases, our primary objectives are threefold:

\begin{itemize}
    \item Our primary objective is to assess the performance of our continuous stochastic modeling framework in capturing the conditional distributions of the closure term under two distinct scenarios. In the first scenario, the closure term exhibits a linear, local dependency on the state, as exemplified by the stochastic viscous diffusion term $G(\mathrm{x}, t)$. In contrast, the second scenario involves a nonlinear, non-local dependency, as illustrated by the stochastic convection term $H(\mathrm{x}, t)$. For each case, we train a separate score-based diffusion model employing Fourier Neural Operators (FNOs) to construct the score functions. These models are conditioned on the vorticity $\omega(\mathrm{x}, t)$ and sparse observations of the true terms $G^\dagger(\mathrm{x}, t)$ and $H^\dagger(\mathrm{x}, t)$. Detailed results for the generated $G$ and $H$ are presented in Sections~\ref{sec: diffusion_generation} and~\ref{sec: nonlinear_generation}, respectively, with additional details on the model architecture and training process provided in Section~\ref{sec: train_detail}.
    
    \item We further investigate and confirm that the continuous stochastic modeling framework is resolution-invariant by training a model based on data with one resolution and testing the model based on data with a few choices of other resolutions. The results are summarized in Section \ref{ssec:results_resolution_invariance}.
    
    \item  We also incorporate the fast sampling technique discussed in Section~\ref{ssec:method_fast_sampling} and employ the trained diffusion models as efficient data-driven closures for $G(\mathrm{x}, t)$ or $H(\mathrm{x}, t)$ in Eq.~\eqref{2dNS}. The accuracy of the simulated vorticity fields and the speedup of the simulations compared to the standard sampling techniques are summarized in Section \ref{sec: surrogate}.
\end{itemize}

To quantitatively evaluate the performance of the generated results, we first use the mean squared error (MSE) defined as
\begin{equation}\label{eqn: MSE}
    D_{\text{MSE}}(G^\dag, G) = \frac{\left \|G^\dag - G\right \|_F^2}{N},
\end{equation}
where $\|\cdot\|_F$ denotes the Frobenius norm and $N$ is the total number of grid points, providing a measure of the average squared error per grid point.

We also adopt a few other relative error metrics in the sense of various matrix norms, hoping to provide a more comprehensive evaluation of our framework's performance. The metrics are defined below:
    \begin{itemize}  
    \item
    Relative Frobenius norm error:
        \begin{equation}\label{eqn: Fro}
            D_{\text{Fro}}(G^\dag, G) = \frac{\left \|G^\dag - G\right\|_F}{\left \| G^\dag\right \|_F},
        \end{equation}
        which quantifies the overall normalized difference between the ground truth $G^\dag$ and the predicted $G$.

    \item
    Relative spectral norm error:
        \begin{equation}\label{eqn: Spe}
            D_{\text{Spe}}(G^\dag, G) = \frac{\left \|G^\dag - G\right\|_2}{\left \| G^\dag\right \|_2},
        \end{equation}
        which measures the largest singular value of the difference that captures the most significant directional error.

    \item
    Relative max norm error:
        \begin{equation}\label{eqn: Max}
            D_{\text{Max}}(G^\dag, G) = \frac{\left \|G^\dag - G\right\|_\text{max}}{\left \| G^\dag\right \|_\text{max}},
        \end{equation}
        which shows the normalized maximum absolute difference between corresponding entries.
    \end{itemize}
Analogous definitions are applied for evaluating the generated convection term $H$ and the simulated vorticity field $\omega$.

To further confirm the resolution invariance property, we also study the energy spectrum between the true fields and the generated ones. The energy spectrum is defined as:
\begin{align}\label{eqn: energy_spectrum}
    E(k, t) = \frac{1}{2} \left |\hat{G}(k, t) \right |^2,
\end{align}
where $\hat{G}(k, t) = \mathcal{F}(G(\mathrm{x}, t))$ denotes the Fourier transform of the field $G(\mathrm{x}, t)$ with $k$ as the wavenumber. $| \cdot |$ evaluates the magnitude of a complex number.

\subsection{Conditional Generation of the Stochastic Viscous Diffusion Term}\label{sec: diffusion_generation}
\begin{figure}[H]
    \centering
    \includegraphics[width = \linewidth]{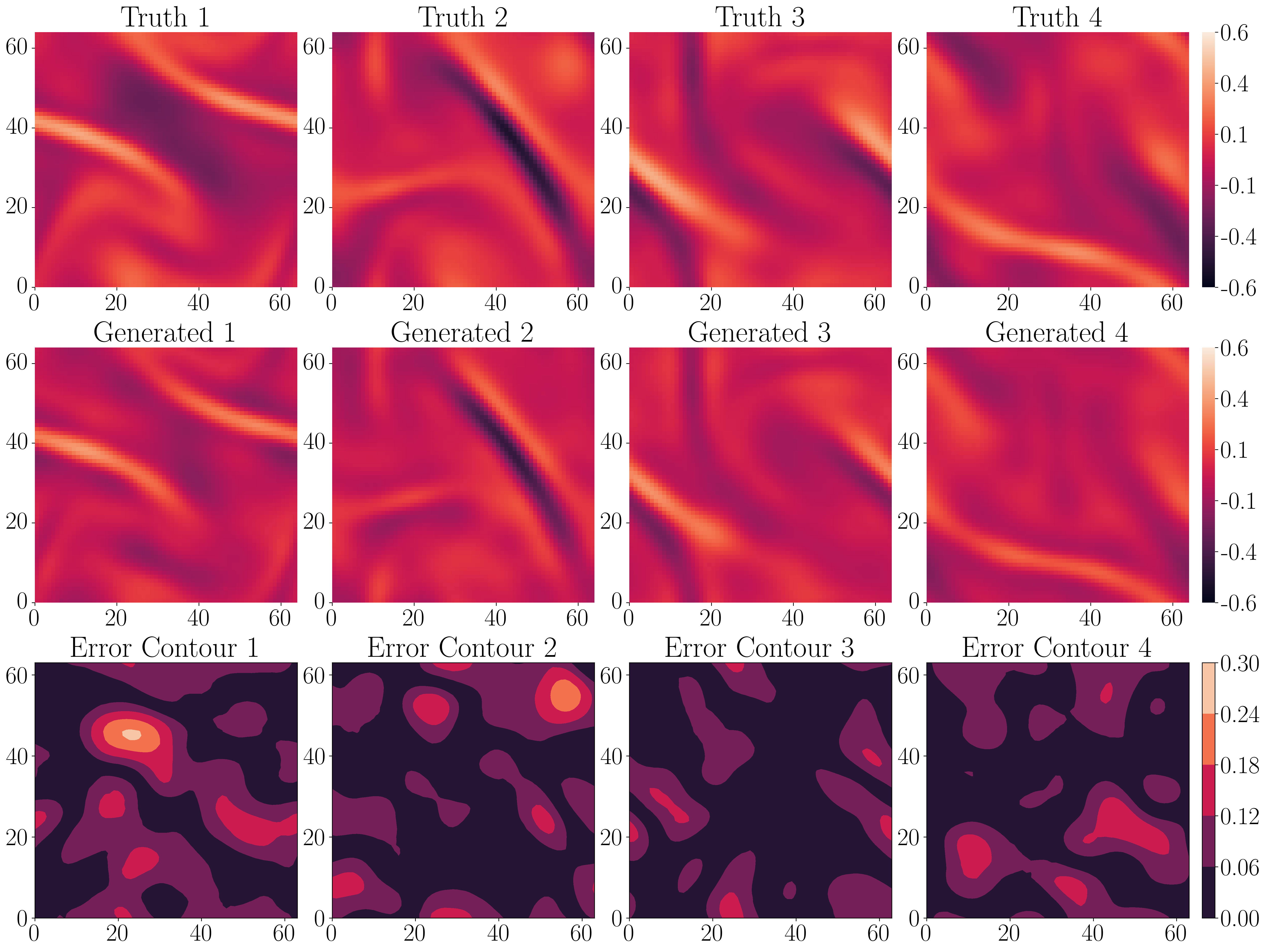}
    \caption{Generated $G(\mathrm{x}, t)$ without sparse information and conditioned only on current vorticity $\omega$. Training data resolution: $64 \times 64$. Test data resolution: $64 \times 64$. First row: random samples of the ground truth $G^\dag$. Second row: corresponding generated samples $G$. Third row: absolute error fields between $G^\dag$ and $G$. The indices $1$ to $4$ correspond to different snapshots randomly sampled from the test dataset.}
    \label{fig: nosparse}
\end{figure}

Based on the proposed diffusion model architecture in Fig.~\ref{fig:schematic}, we first train a conditional diffusion model for the stochastic viscous diffusion term $G^\dag(\mathrm{x}, t)$ in Eq.~\eqref{eqn: diffusion}. Figure~\ref{fig: nosparse} presents the test data results of the stochastic viscous diffusion terms of the true system and the generated ones, which are based on the trained conditional diffusion model that is only conditioned on the vorticity field $\omega$ at the current time. As shown in Fig.~\ref{fig: nosparse}, the discrepancies between the ground truth $G^\dag(\mathrm{x}, t)$ and the generated samples $G(\mathrm{x}, t)$ remain significant when the diffusion model is conditioned solely on the current vorticity $\omega(\mathrm{x}, t)$. Across all batches, the mean squared error ($D_{\text{MSE}}$) is $5.4024 \times 10^{-3}$, while the average relative Frobenius norm ($D_{\text{Fro}}$), spectral norm ($D_{\text{Spe}}$), and max norm ($D_{\text{Max}}$) are  $5.9448 \times 10^{-1}$, $7.3514\times 10^{-1}$, and $5.2182\times 10^{-1}$ respectively.

To enhance the model's capability in capturing the information in the ground truth $G^\dag(\mathrm{x}, t)$, we further incorporate sparse observations of $G^\dag(\mathrm{x}, t)$ as an extra conditional input into the current model setup. Given sparse observation vectors $G^\dagger_{\textrm{sparse}}$ (with a resolution of $16 \times 16$ in this work), multiple methods are available for upscaling the sparse measurements to match the resolution of the model ($64 \times 64$), including interpolation and convolution. However, as shown in Table~\ref{tab: traditional_upscale}, these methods alone are insufficient for accurately reconstructing the field based on sparse measurements. The upscaling of sparse measurements is achieved through two commonly used techniques: (i) bicubic interpolation and (ii) 2-D convolution.

\begin{table}[H]
\centering
\caption{Errors of upscaled sparse observation of the ground truth $G^\dagger_{\textrm{sparse}}$ from the resolution of $16 \times 16$ to the one of $64 \times 64$ via bicubic interpolation and 2-D convolution.}
\begin{tabular}{|c|cc|cccc|}
\hline
\multirow{2}{*}{\diagbox[width=11em,height=2.5em,dir=SE]{Methods}{Errors}} & \multicolumn{2}{c|}{Test Data}                    & \multicolumn{4}{c|}{Test Errors} \\ \cline{2-7} 
                  & \multicolumn{1}{c|}{$\mathrm{d}x$} & $\mathrm{d}y$ & \multicolumn{1}{c|}{$D_{\text{MSE}}$}           & \multicolumn{1}{c|}{$D_{\text{Fro}}$} 
                  & \multicolumn{1}{c|}{$D_{\text{Spe}}$}   & \multicolumn{1}{c|}{$D_{\text{Max}}$}   \\ \hline
                  
Bicubic interpolation      & \multicolumn{1}{c|}{1/64}          & 1/64         & \multicolumn{1}{c|}{1.2714e-3}      & \multicolumn{1}{c|}{2.9726e-1} & \multicolumn{1}{c|}{2.7906e-1} & \multicolumn{1}{c|}{5.7404e-1}\\ \hline

2-D convolution & \multicolumn{1}{c|}{1/64}          & 1/64         & \multicolumn{1}{c|}{3.4697e-3}      & \multicolumn{1}{c|}{4.9590e-1} & \multicolumn{1}{c|}{4.8228e-1}  & \multicolumn{1}{c|}{7.0584e-1}  \\ \hline
\end{tabular}

\label{tab: traditional_upscale}
\end{table}

\begin{figure}[H]
    \centering
    \includegraphics[width = \linewidth]{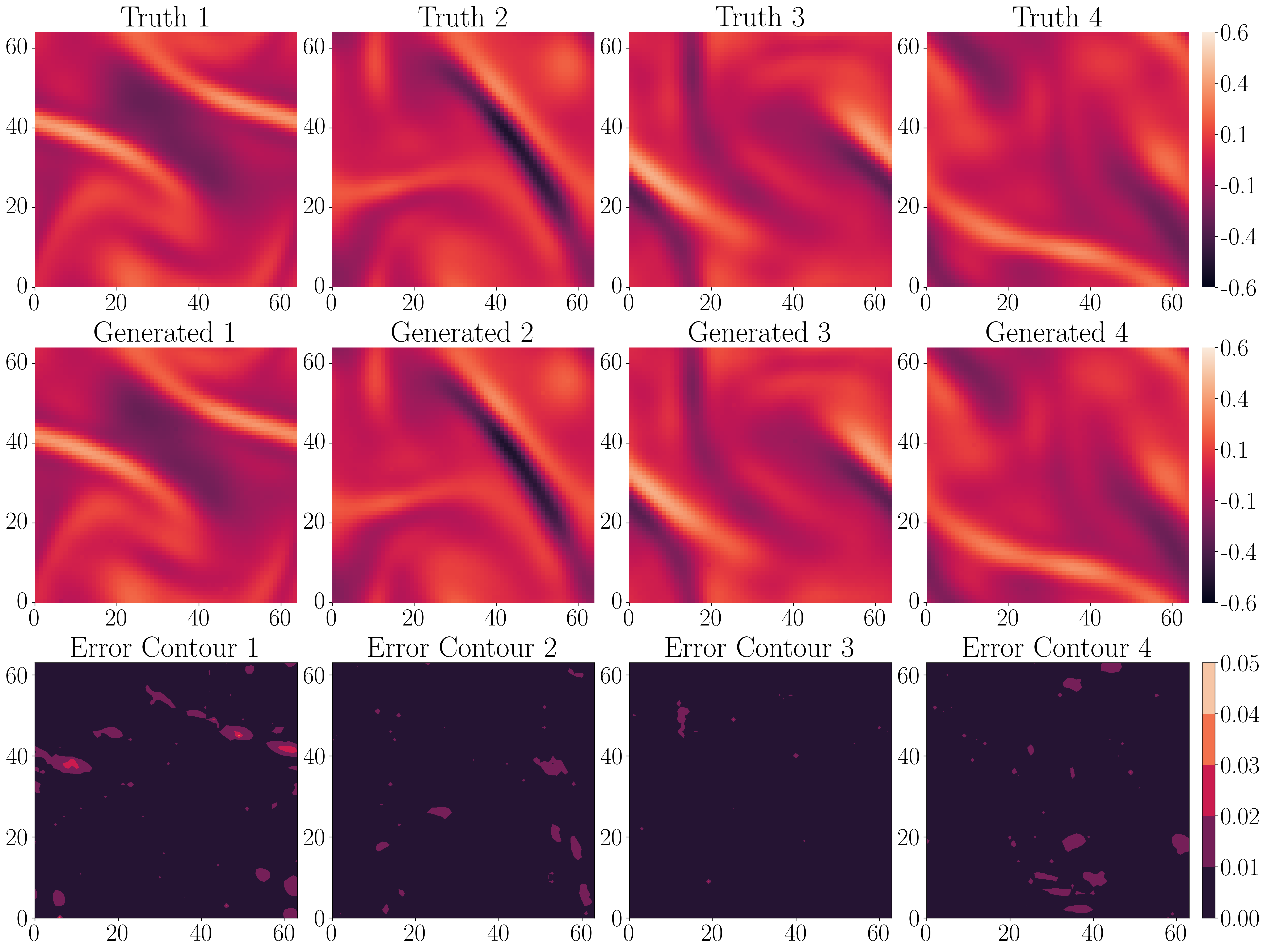}
    \caption{Generated $G(\mathrm{x},t)$ conditioned on current vorticity $\omega$ and sparse information of $G^\dagger_{\textrm{sparse}}$. Sparse information is upscaled using bicubic interpolation. Training data resolution: $64 \times 64$. Test data resolution: $64 \times 64$. First row: random samples of the ground truth $G^\dag$. Second row: corresponding generated samples $G$. Third row: absolute error fields between the ground truth $G^\dag$ and generated $G$.}
    \label{fig:G_test64_interp}
\end{figure}

\begin{figure}[H]
    \centering
    \includegraphics[width = \linewidth]{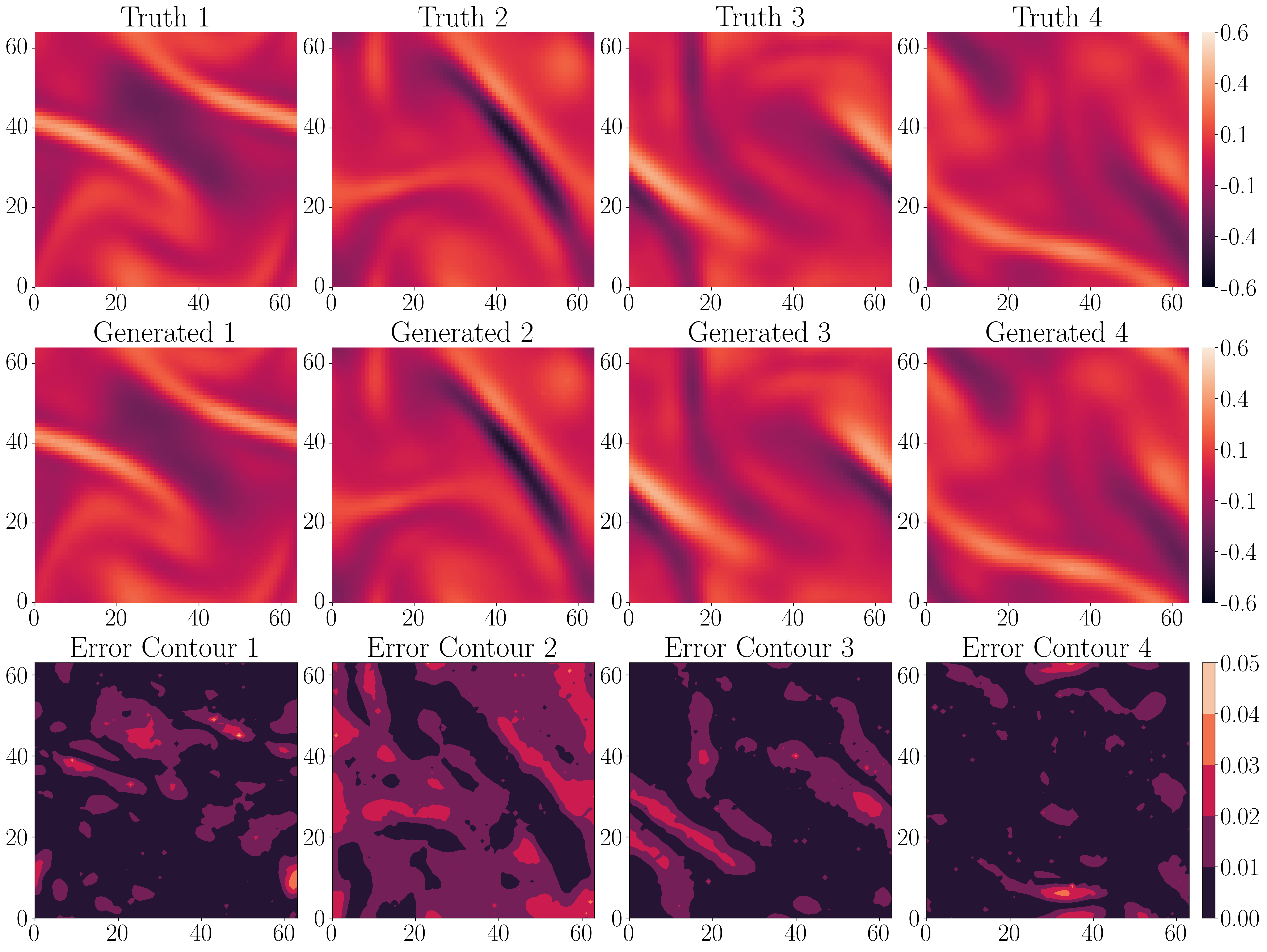}
    \caption{Generated $G(\mathrm{x},t)$ conditioned on current vorticity $\omega$ and sparse information of $G^\dagger_{\textrm{sparse}}$. Sparse information is upscaled using 2-D convolution. Training data resolution: $64 \times 64$. Test data resolution: $64 \times 64$. First row: random samples of the ground truth $G^\dag$. Second row: corresponding generated samples $G$. Third row: absolute error fields between the ground truth $G^\dag$ and generated $G$.}
    \label{fig:G_test64_conv}
\end{figure}

On the other hand, we demonstrate that incorporating upscaled sparse measurements as an additional input condition of the proposed conditional diffusion model effectively enhances the model's performance in reconstructing the ground truth $G^\dag(\mathrm{x}, t)$. The improvements in the quality of the generated samples are confirmed by Fig.~\ref{fig:G_test64_interp} and Fig.~\ref{fig:G_test64_conv}, both of which provide smaller absolute error fields compared to the ones in Fig.~\ref{fig: nosparse}. As summarized in Table~\ref{tab: with_sparse}, the conditional diffusion model without the sparse information of $G^\dag$ struggles to provide a reliable generative model for the stochastic viscous diffusion term, resulting in high relative errors. However, when the model is conditioned on both vorticity $\omega(\mathrm{x}, t)$ and upscaled sparse information of $G^\dag_{\text{sparse}}(\mathrm{x}, t)$, the accuracy of the generated samples significantly improves. The main reason is that conditioning on sparse information of $G^\dag$ avoids the need to accurately train the score function within the whole high-dimensional probability space and only focuses on those highly correlated regimes with respect to the incorporated sparse information of $G^\dag$, and thus better performance can be expected with a limited amount of data. Meanwhile, since the interpolated sparse observations already achieve a better agreement with the ground truth $G^\dag(\mathrm{x}, t)$, compared to those upscaled using 2-D convolution techniques, it is expected that our model can achieve a better performance when conditioned on the interpolated $G^\dag_{\text{sparse}}(\mathrm{x}, t)$.

\begin{table}[H]
\centering
\caption{Errors of the generated $G$ for test data with a resolution of $64 \times 64$. The first row presents the errors when the model is not conditioned on any sparse information $G^\dagger_{\textrm{sparse}}$. The second row presents the errors when the model is conditioned on bicubically interpolated sparse observations. The third row presents the errors when the model is conditioned on sparsely observed data processed with 2-D convolution.}
\begin{tabular}{|c|cc|cccc|}
\hline
\multirow{2}{*}{\diagbox[dir=SE]{Methods}{Errors}} & \multicolumn{2}{c|}{Test Data}                    & \multicolumn{4}{c|}{Test Errors} \\ \cline{2-7} 
                  & \multicolumn{1}{c|}{$\mathrm{d}x$} & $\mathrm{d}y$ & \multicolumn{1}{c|}{$D_{\text{MSE}}$}            & \multicolumn{1}{c|}{$D_{\text{Fro}}$} & \multicolumn{1}{c|}{$D_{\text{Spe}}$} & \multicolumn{1}{c|}{$D_{\text{Max}}$}   \\ \hline
No sparse & \multicolumn{1}{c|}{1/64}          & 1/64         & \multicolumn{1}{c|}{5.4024e-3}    & \multicolumn{1}{c|}{5.9448e-1} & \multicolumn{1}{c|}{7.3514e-1} & \multicolumn{1}{c|}{5.2182e-1}   \\ \hline

Interpolation      & \multicolumn{1}{c|}{1/64}          & 1/64         & \multicolumn{1}{c|}{2.3340e-5}    & \multicolumn{1}{c|}{3.7137e-2} & \multicolumn{1}{c|}{3.6080e-2} & \multicolumn{1}{c|}{7.0687e-2}  \\ \hline

Convolution & \multicolumn{1}{c|}{1/64}          & 1/64         & \multicolumn{1}{c|}{1.4778e-4}    & \multicolumn{1}{c|}{8.8991e-2} & \multicolumn{1}{c|}{1.1519e-1} & \multicolumn{1}{c|}{1.0383e-1}  \\ \hline
\end{tabular}

\label{tab: with_sparse}
\end{table}

\subsection{Conditional Generation of the Stochastic Convection Term}\label{sec: nonlinear_generation}
\begin{figure}[H]
    \centering
    \includegraphics[width = \linewidth]{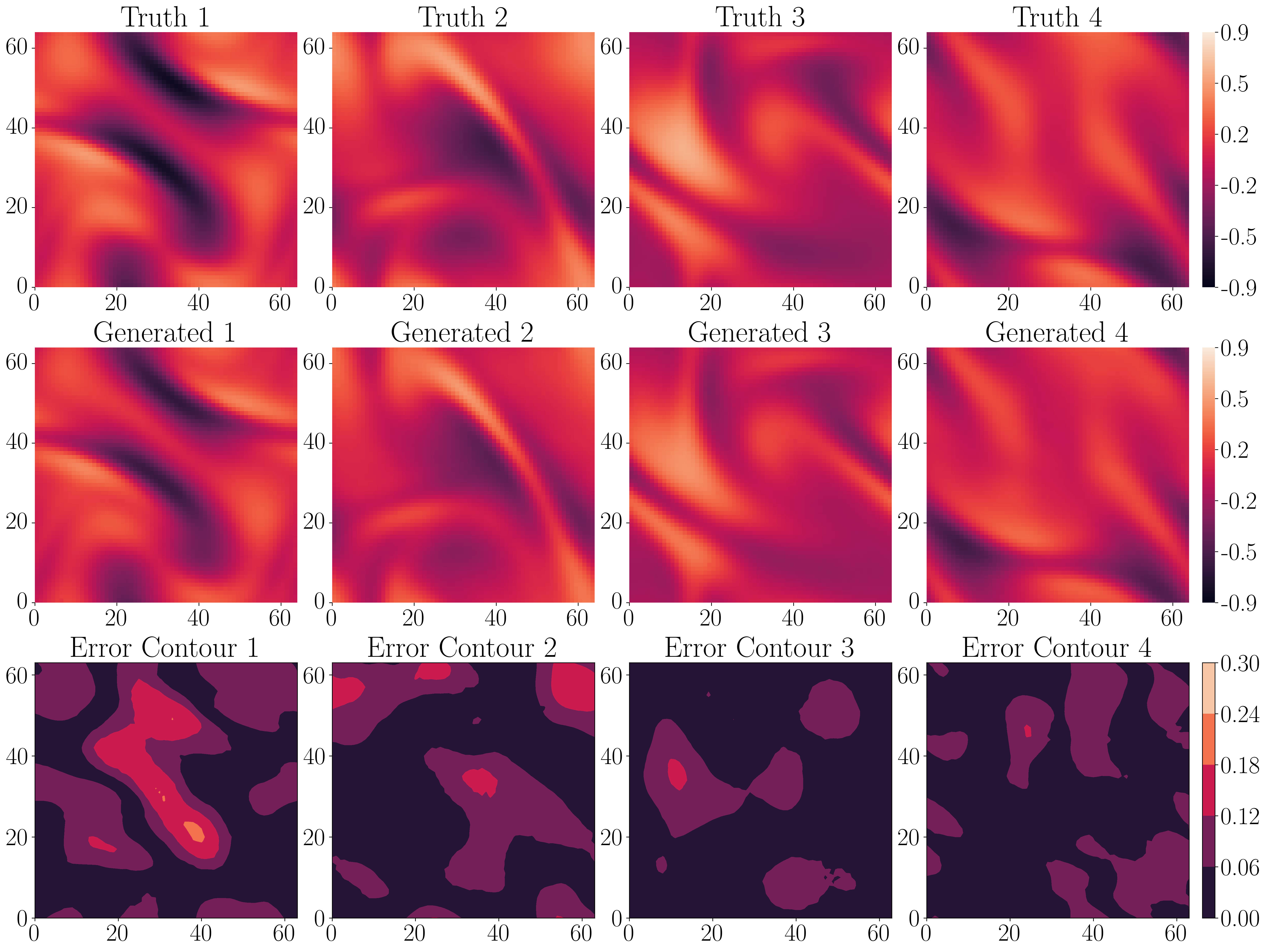}
    \caption{Generated $H(\mathrm{x}, t)$ without sparse information and conditioned only on current vorticity $\omega$. Training data resolution: $64 \times 64$. Test data resolution: $64 \times 64$. First row: random samples of the ground truth $H^\dag$. Second row: corresponding generated samples $H$. Third row: absolute error fields between $H^\dag$ and $H$. The indices $1$ to $4$ correspond to different snapshots randomly sampled from the test dataset.}
    \label{fig: nosparse_H}
\end{figure}

We now focus on the nonlinear and non-local dependency scenario by training a conditional diffusion model to capture the stochastic convection term \( H^\dagger(\mathrm{x}, t) \) as defined in Eq.~\eqref{eqn: convection}. Figure~\ref{fig: nosparse_H} presents the ground truth $H^\dag(\mathrm{x}, t)$ and the generated samples that are produced by the trained conditional diffusion model, which remains conditioned exclusively on the vorticity field $\omega$ of the current time step. As shown in Fig.~\ref{fig: nosparse_H}, significant discrepancies persist between the ground truth and the generated samples when the model is conditioned solely on the current vorticity $\omega(\mathrm{x}, t)$. Quantitatively, $D_{\text{MSE}}$ is $3.9311 \times 10^{-3}$, while $D_{\text{Fro}}$, $D_{\text{Spe}}$, and $D_{\text{Max}}$ are  $2.6164 \times 10^{-1}$, $2.9418 \times 10^{-1}$, and $2.5535 \times 10^{-1}$ respectively. It can be noted that these discrepancies are notably smaller than those observed for the stochastic viscous diffusion term. This improvement is attributed to the larger magnitude of the convection terms, which diminishes the relative impact of stochastic fluctuations.

Following the approach outlined in Section~\ref{sec: diffusion_generation}, we will once again integrate sparse observations of $H^\dag(\mathrm{x}, t)$ as an additional conditional input into the current model framework. These sparse observations, represented as vectors $H^\dagger_{\textrm{sparse}}$ with a resolution of $16 \times 16$ in this part, are upscaled using bicubic interpolation and 2-D convolution to match the model's resolution of $64 \times 64$. We first show in Table~\ref{tab: traditional_upscale_H} that these traditional methods alone are still insufficient for accurately reconstructing the field.

\begin{table}[H]
\centering
\caption{Errors of upscaled sparse observation of the ground truth $H^\dagger_{\textrm{sparse}}$ from the resolution of $16 \times 16$ to the one of $64 \times 64$ via bicubic interpolation and 2-D convolution.}
\begin{tabular}{|c|cc|cccc|}
\hline
\multirow{2}{*}{\diagbox[width=11em,height=2.5em,dir=SE]{Methods}{Errors}} & \multicolumn{2}{c|}{Test Data}                    & \multicolumn{4}{c|}{Test Errors} \\ \cline{2-7} 
                  & \multicolumn{1}{c|}{$\mathrm{d}x$} & $\mathrm{d}y$ & \multicolumn{1}{c|}{$D_{\text{MSE}}$}           & \multicolumn{1}{c|}{$D_{\text{Fro}}$} 
                  & \multicolumn{1}{c|}{$D_{\text{Spe}}$}   & \multicolumn{1}{c|}{$D_{\text{Max}}$}   \\ \hline
                  
Bicubic interpolation      & \multicolumn{1}{c|}{1/64}          & 1/64         & \multicolumn{1}{c|}{3.3143e-3}      & \multicolumn{1}{c|}{2.3327e-1} & \multicolumn{1}{c|}{1.9937e-1} & \multicolumn{1}{c|}{5.0485e-1}\\ \hline

2-D convolution & \multicolumn{1}{c|}{1/64}          & 1/64         & \multicolumn{1}{c|}{1.0641e-2}      & \multicolumn{1}{c|}{4.2516e-1} & \multicolumn{1}{c|}{3.6728e-1}  & \multicolumn{1}{c|}{7.2317e-1}  \\ \hline
\end{tabular}

\label{tab: traditional_upscale_H}
\end{table}

\begin{figure}[H]
    \centering
    \includegraphics[width = \linewidth]{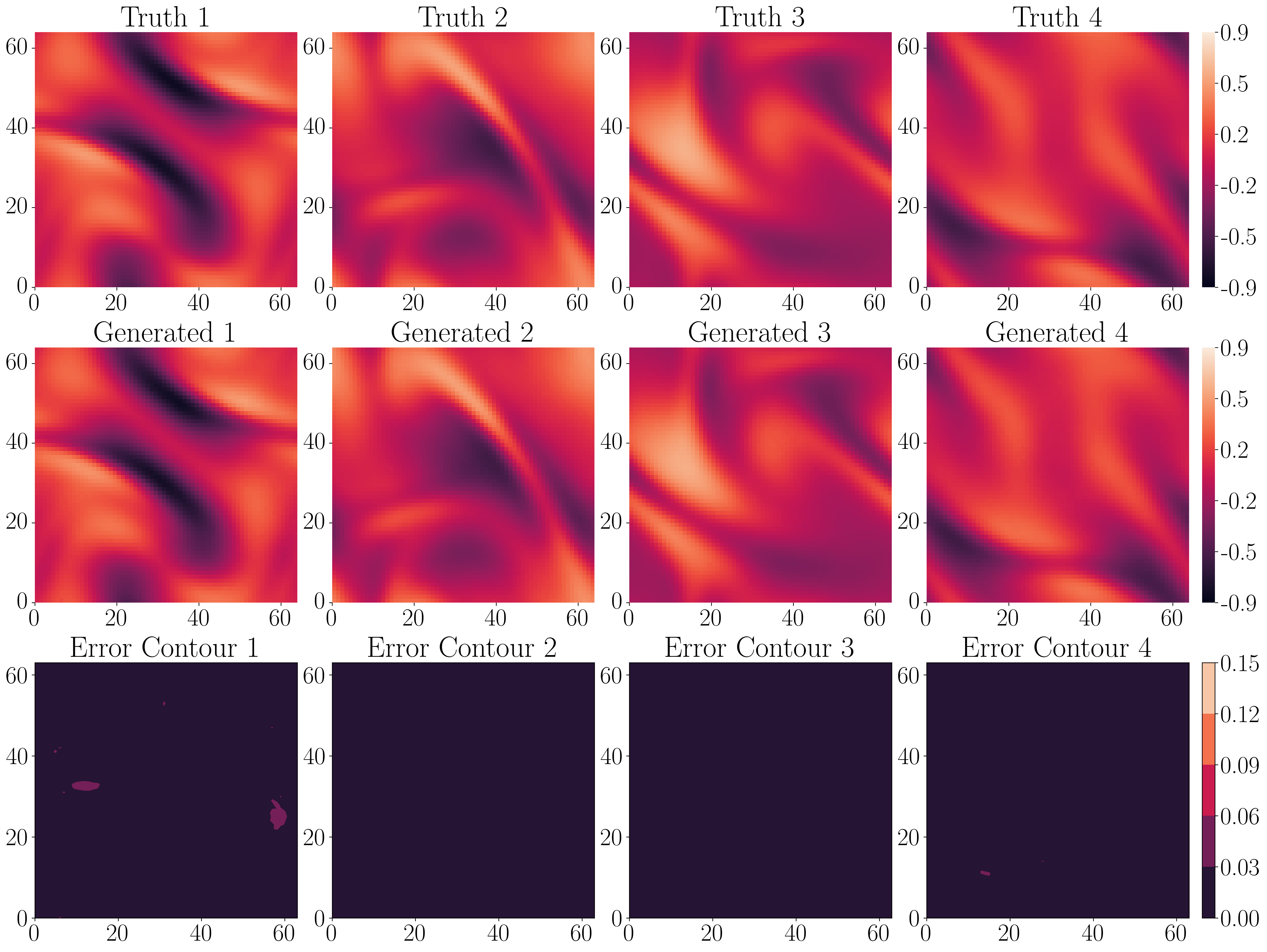}
    \caption{Generated $H(\mathrm{x},t)$ conditioned on current vorticity $\omega$ and sparse information of $H^\dagger_{\textrm{sparse}}$. Sparse information is upscaled using bicubic interpolation. Training data resolution: $64 \times 64$. Test data resolution: $64 \times 64$. First row: random samples of the ground truth $H^\dag$. Second row: corresponding generated samples $H$. Third row: absolute error fields between the ground truth $H^\dag$ and generated $H$.}
    \label{fig:H_test64_interp}
\end{figure}

\begin{figure}[H]
    \centering
    \includegraphics[width = \linewidth]{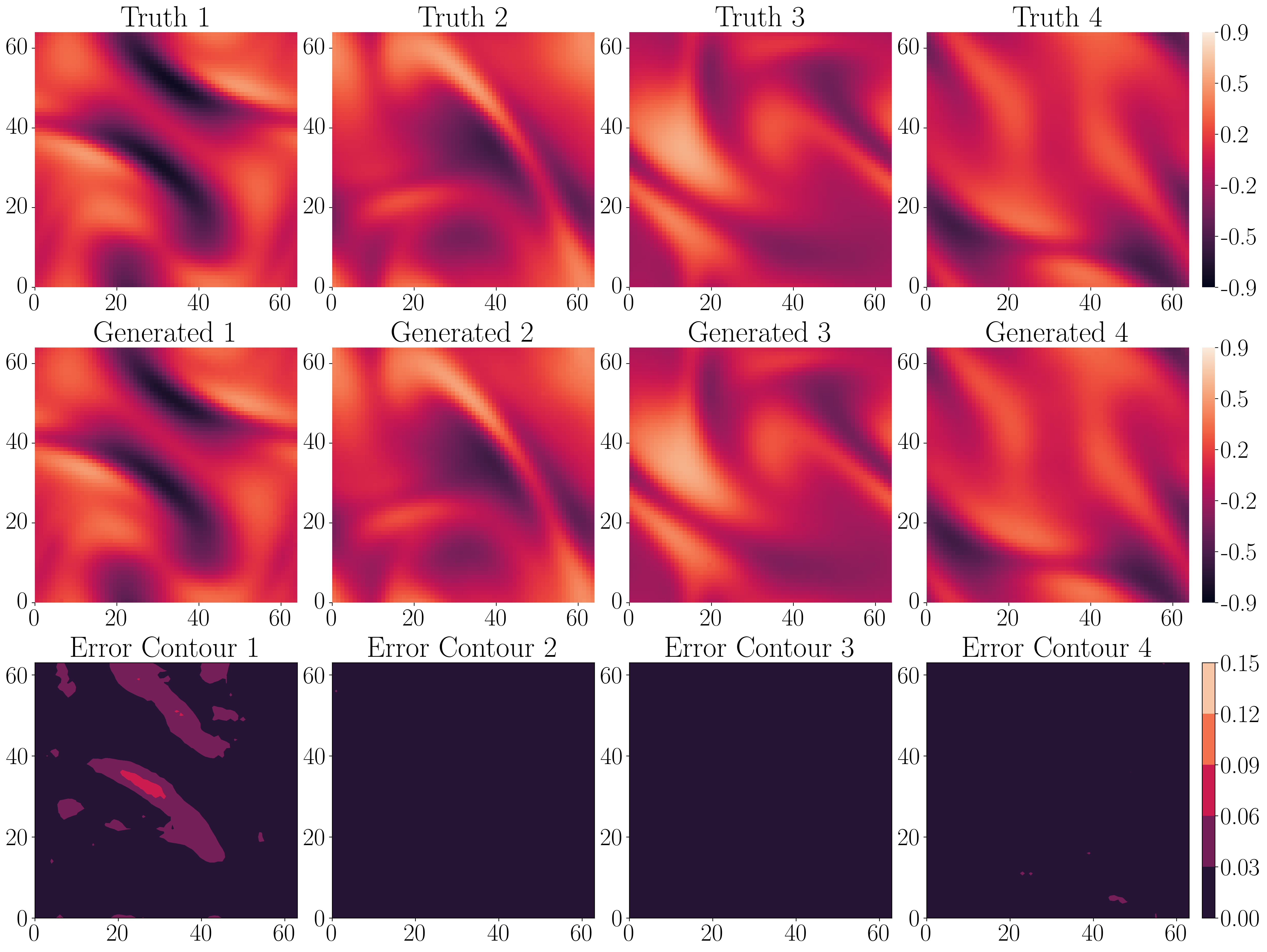}
    \caption{Generated $H(\mathrm{x},t)$ conditioned on current vorticity $\omega$ and sparse information of $H^\dagger_{\textrm{sparse}}$. Sparse information is upscaled using 2-D convolution. Training data resolution: $64 \times 64$. Test data resolution: $64 \times 64$. First row: random samples of the ground truth $H^\dag$. Second row: corresponding generated samples $H$. Third row: absolute error fields between the ground truth $H^\dag$ and generated $H$.}
    \label{fig:H_test64_conv}
\end{figure}

The improvements in sample quality after incorporating sparse information are demonstrated in Figs.~\ref{fig:H_test64_interp} and~\ref{fig:H_test64_conv}, which show notably smaller absolute error fields compared to Fig.~\ref{fig: nosparse_H}. As detailed in Table~\ref{tab: with_sparse_H}, a conditional diffusion model that lacks the sparse representation of \(H^\dag\) struggles to accurately generate the stochastic nonlinear advection term, resulting in high relative errors. In contrast, conditioning the model on both the vorticity \(\omega(\mathrm{x}, t)\) and the upscaled sparse information \(H^\dag_{\text{sparse}}(\mathrm{x}, t)\) leads to a significant enhancement in the accuracy of the generated samples. Furthermore, when the upscaled sparse data more closely matches the ground truth \(H^\dag(\mathrm{x}, t)\), its incorporation as an additional conditioning input further boosts the model's performance.

\begin{table}[H]
\centering
\caption{Errors of the generated $H$ for test data with a resolution of $64 \times 64$. The first row presents the errors when the model is not conditioned on any sparse information $H^\dagger_{\textrm{sparse}}$. The second row presents the errors when the model is conditioned on bicubically interpolated sparse observations. The third row presents the errors when the model is conditioned on sparsely observed data processed with 2-D convolution.}
\begin{tabular}{|c|cc|cccc|}
\hline
\multirow{2}{*}{\diagbox[dir=SE]{Methods}{Errors}} & \multicolumn{2}{c|}{Test Data}                    & \multicolumn{4}{c|}{Test Errors} \\ \cline{2-7} 
                  & \multicolumn{1}{c|}{$\mathrm{d}x$} & $\mathrm{d}y$ & \multicolumn{1}{c|}{$D_{\text{MSE}}$}            & \multicolumn{1}{c|}{$D_{\text{Fro}}$} & \multicolumn{1}{c|}{$D_{\text{Spe}}$} & \multicolumn{1}{c|}{$D_{\text{Max}}$}   \\ \hline
No sparse & \multicolumn{1}{c|}{1/64}          & 1/64         & \multicolumn{1}{c|}{3.9311e-3}    & \multicolumn{1}{c|}{2.6164e-1} & \multicolumn{1}{c|}{2.9418e-1} & \multicolumn{1}{c|}{2.5535e-1}   \\ \hline

Interpolation      & \multicolumn{1}{c|}{1/64}          & 1/64         & \multicolumn{1}{c|}{1.0850e-4}    & \multicolumn{1}{c|}{3.8475e-2} & \multicolumn{1}{c|}{3.7667e-2} & \multicolumn{1}{c|}{5.6713e-2}  \\ \hline

Convolution & \multicolumn{1}{c|}{1/64}          & 1/64         & \multicolumn{1}{c|}{2.3878e-4}    & \multicolumn{1}{c|}{5.5936e-2} & \multicolumn{1}{c|}{5.6179e-2} & \multicolumn{1}{c|}{7.9217e-2}  \\ \hline
\end{tabular}

\label{tab: with_sparse_H}
\end{table}

\subsection{Resolution Invariance of the Trained Models}
\label{ssec:results_resolution_invariance}

\begin{figure}[H]
    \centering
    \includegraphics[width = \linewidth]{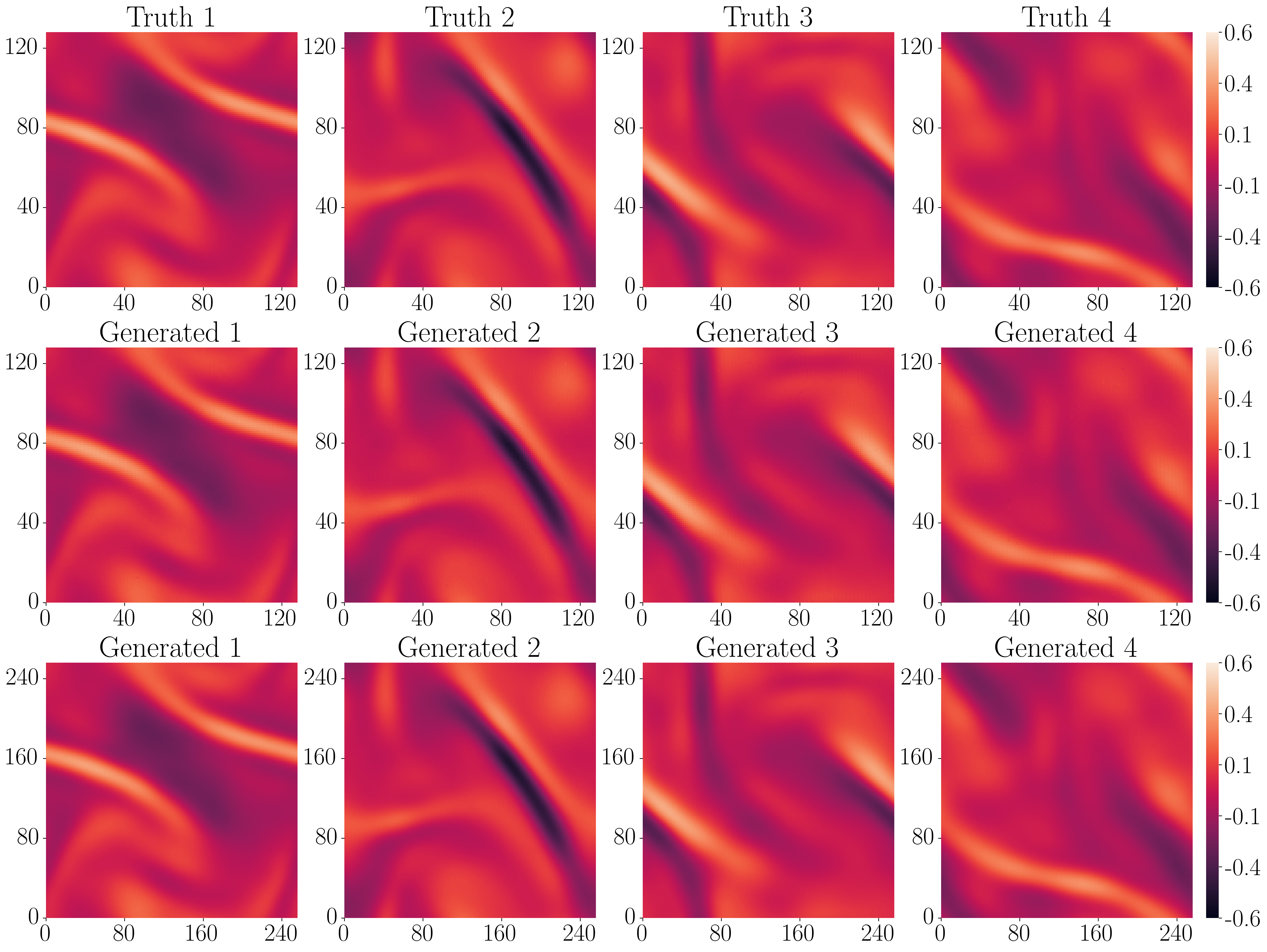}
    \caption{Generated $G(\mathrm{x},t)$ conditioned on current vorticity $\omega$ and sparse information of $G^\dagger_{\textrm{sparse}}$. Sparse information is upscaled using bicubic interpolation. Training data resolution: $64 \times 64$. Test data resolution: $128 \times 128$ and $256 \times 256$. First row: random samples of the ground truth $\left.G^\dag\right|_{128 \times 128}$. The ground truth $\left.G^\dag\right|_{256 \times 256}$ is visually identical hence omitted. Second row: corresponding generated samples $\left.G\right|_{128 \times 128}$. Third row: corresponding generated samples $\left.G\right|_{256 \times 256}$.}
    \label{fig:G_test_interp}
\end{figure}

\begin{figure}[H]
    \centering
    \includegraphics[width = \linewidth]{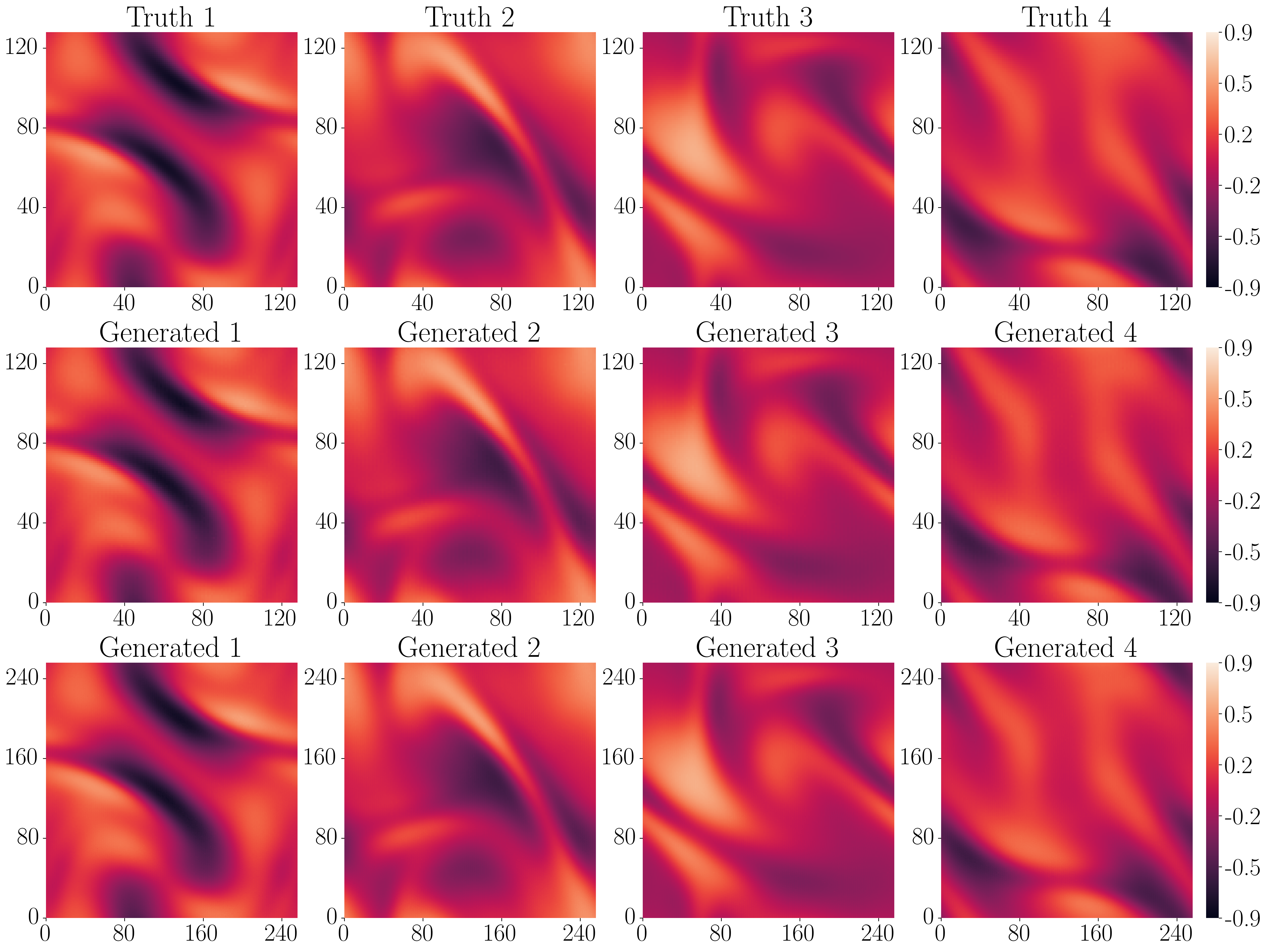}
    \caption{Generated $H(\mathrm{x},t)$ conditioned on current vorticity $\omega$ and sparse information of $H^\dagger_{\textrm{sparse}}$. Sparse information is upscaled using bicubic interpolation. Training data resolution: $64 \times 64$. Test data resolution: $128 \times 128$ and $256 \times 256$. First row: random samples of the ground truth $\left.H^\dag\right|_{128 \times 128}$. The ground truth $\left.H^\dag\right|_{256 \times 256}$ is visually identical hence omitted. Second row: corresponding generated samples $\left.H\right|_{128 \times 128}$. Third row: corresponding generated samples $\left.H\right|_{256 \times 256}$.}
    \label{fig:H_test_interp}
\end{figure}

In this section, we demonstrate the resolution invariance of the proposed stochastic modeling framework. Figures~\ref{fig:G_test_interp} and~\ref{fig:H_test_interp} compare the conditionally generated closure terms with the ground truth. In both cases—whether modeling \( G \) or \( H \)—the models are evaluated on finer resolutions (\(128 \times 128\) and \(256 \times 256\)) that differ from the training resolution (\(64 \times 64\)). Sparse observations of \( G^\dagger \) and \( H^\dagger \) at \(16 \times 16\) are incorporated via bicubic interpolation. Notably, the use of FNOs unlocks super-resolution capabilities, enabling the model to infer closure terms at resolutions finer than those seen during training. As shown in Fig.~\ref{fig:G_test_interp} and Fig.~\ref{fig:H_test_interp}, the generated results consistently capture the overall patterns of the true system, at least on a qualitative level.

\begin{figure}[H]
    \centering
    \includegraphics[width = \linewidth]{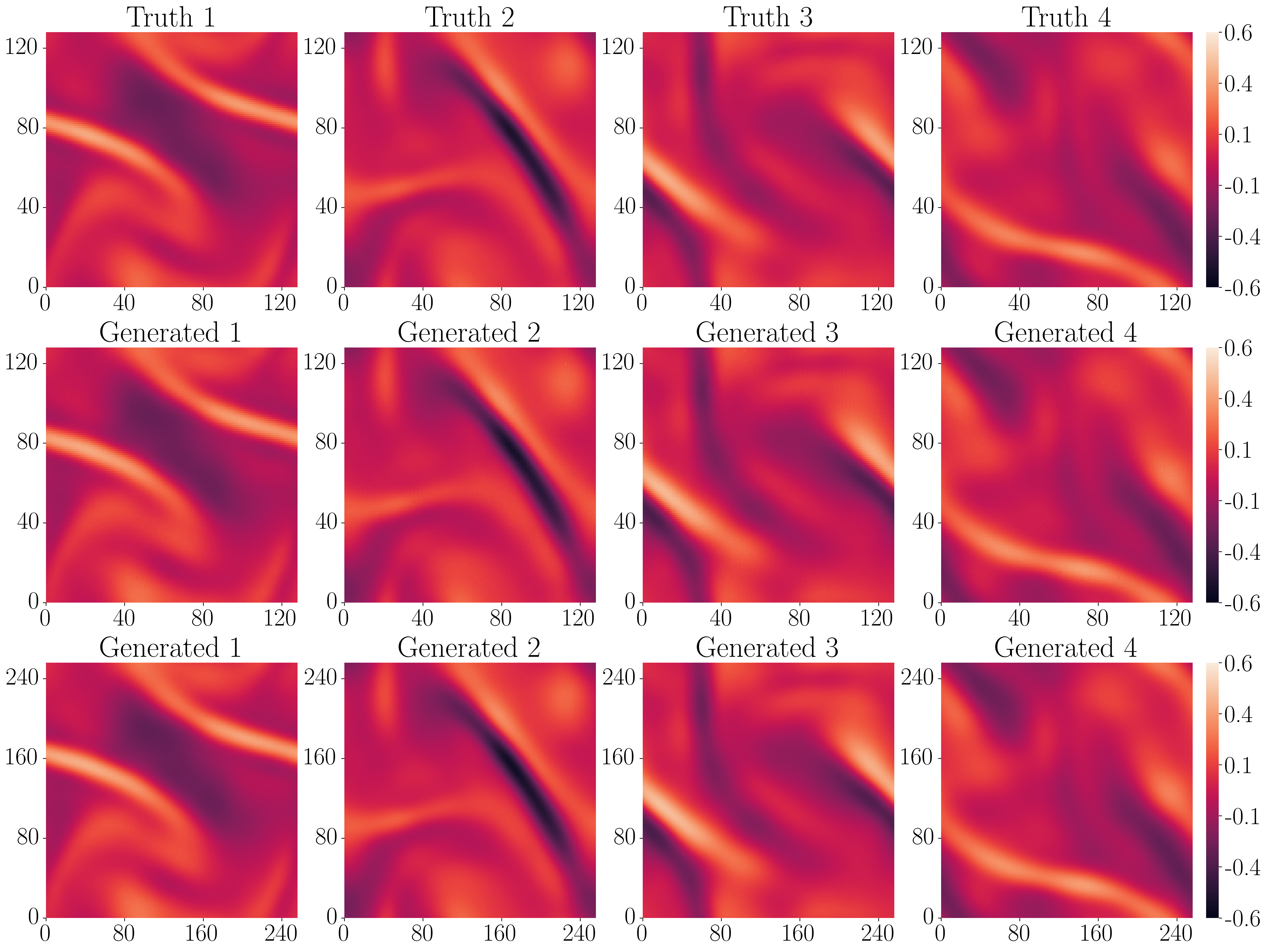}
    \caption{Generated $G(\mathrm{x},t)$ conditioned on current vorticity $\omega$ and sparse information of $G^\dagger_{\textrm{sparse}}$. Sparse information is upscaled using 2-D convolution. Training data resolution: $64 \times 64$. Test data resolution: $128 \times 128$ and $256 \times 256$. First row: random samples of the ground truth $\left.G^\dag\right|_{128 \times 128}$. The ground truth $\left.G^\dag\right|_{256 \times 256}$ is visually identical hence omitted. Second row: corresponding generated samples $\left.G\right|_{128 \times 128}$. Third row: corresponding generated samples $\left.G\right|_{256 \times 256}$.}
    \label{fig:G_test_conv}
\end{figure}

\begin{figure}[H]
    \centering
    \includegraphics[width = \linewidth]{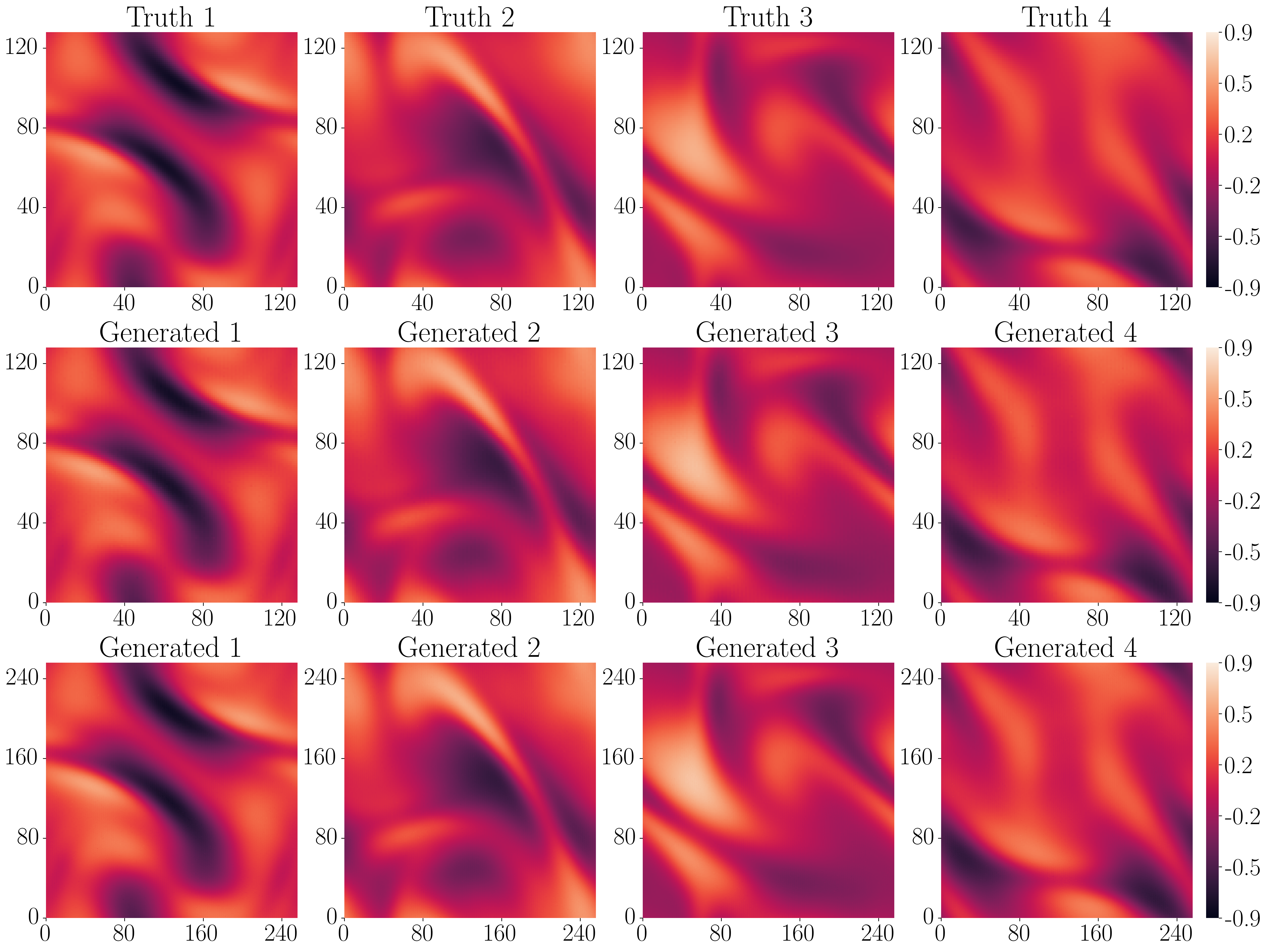}
    \caption{Generated $H(\mathrm{x},t)$ conditioned on current vorticity $\omega$ and sparse information of $H^\dagger_{\textrm{sparse}}$. Sparse information is upscaled using 2-D convolution. Training data resolution: $64 \times 64$. Test data resolution: $128 \times 128$ and $256 \times 256$. First row: random samples of the ground truth $\left.H^\dag\right|_{128 \times 128}$. The ground truth $\left.H^\dag\right|_{256 \times 256}$ is visually identical hence omitted. Second row: corresponding generated samples $\left.H\right|_{128 \times 128}$. Third row: corresponding generated samples $\left.H\right|_{256 \times 256}$.}
    \label{fig:H_test_conv}
\end{figure}

In Fig.~\ref{fig:G_test_conv} and Fig.~\ref{fig:H_test_conv}, we present results of utilizing 2-D convolution to preprocess sparse information. The kernel size of the 2-D convolution layers to preprocess the sparse information scales with different target resolutions, i.e., for a target of $64 \times 64$, we use a kernel size of 7, while for $128 \times 128$ and $256 \times 256$, we use 15 and 31 respectively. Similar to the results in Fig.~\ref{fig:G_test_interp} and Fig.~\ref{fig:H_test_interp}, it can be seen in Fig.~\ref{fig:G_test_conv} and Fig.~\ref{fig:H_test_conv} that the test results at different resolutions show a good agreement with the true system, which demonstrates that the resolution invariance property of the trained models is not sensitive to the specific technique for upscaling the sparse conditional information.

In Table~\ref{tab: resolution_invariance} and Table~\ref{tab: resolution_invariance_H}, we present quantitative results that further confirm the resolution invariance of our model. Recall that all models were trained at a resolution of $64 \times 64$ and evaluated at different resolutions ($64 \times 64$, $128 \times 128$, and $256 \times 256$). The results show no significant drop in performance when tested at finer resolutions, demonstrating the robustness of our approach across varying input scales. This resolution-invariance property primarily stems from the use of FNOs to construct the score function with multimodal inputs.

\begin{table}[H]
\centering
\caption{Errors of the generated $G$ for test data with different resolutions of $64 \times 64$, $128 \times 128$, and $256 \times 256$. The training resolution is $64 \times 64$. The first to third rows present the errors when the model is conditioned on bicubically interpolated sparse observations, whereas the fourth to sixth rows present the errors when the model is conditioned on sparsely observed data processed with 2-D convolution.}
\begin{tabular}{|c|cc|cccc|}
\hline
\multirow{2}{*}{\diagbox[dir=SE]{Methods}{Errors}} & \multicolumn{2}{c|}{Test Data}                    & \multicolumn{4}{c|}{Test Errors} \\ \cline{2-7} 
                  & \multicolumn{1}{c|}{$\mathrm{d}x$} & $\mathrm{d}y$ & \multicolumn{1}{c|}{$D_{\text{MSE}}$}            & \multicolumn{1}{c|}{$D_{\text{Fro}}$} & \multicolumn{1}{c|}{$D_{\text{Spe}}$} & \multicolumn{1}{c|}{$D_{\text{Max}}$}   \\ \hline
                  
\multirow{3}{*}{Interpolation} & \multicolumn{1}{c|}{1/64}          & 1/64           & \multicolumn{1}{c|}{2.3340e-5}    & \multicolumn{1}{c|}{3.7137e-2} & \multicolumn{1}{c|}{3.6080e-2} & \multicolumn{1}{c|}{7.0687e-2} \\ \cline{2-7} 
                               & \multicolumn{1}{c|}{1/128}         & 1/128         & \multicolumn{1}{c|}{5.8047e-5} & \multicolumn{1}{c|}{5.9622e-2} & \multicolumn{1}{c|}{5.7210e-2} & \multicolumn{1}{c|}{1.0416e-1} \\ \cline{2-7} 
                               & \multicolumn{1}{c|}{1/256}         & 1/256         & \multicolumn{1}{c|}{9.5129e-5} & \multicolumn{1}{c|}{7.7193e-2} & \multicolumn{1}{c|}{7.7055e-2} & \multicolumn{1}{c|}{1.4548e-1} \\ \hline
\multirow{3}{*}{Convolution}   & \multicolumn{1}{c|}{1/64}          & 1/64            & \multicolumn{1}{c|}{1.4778e-4}    & \multicolumn{1}{c|}{8.8991e-2} & \multicolumn{1}{c|}{1.1519e-1} & \multicolumn{1}{c|}{1.0383e-1} \\ \cline{2-7} 
                               & \multicolumn{1}{c|}{1/128}         & 1/128         & \multicolumn{1}{c|}{2.2941e-4} & \multicolumn{1}{c|}{1.2112e-1} & \multicolumn{1}{c|}{1.4289e-1} & \multicolumn{1}{c|}{1.3380e-1} \\ \cline{2-7} 
                               & \multicolumn{1}{c|}{1/256}         & 1/256         & \multicolumn{1}{c|}{3.5763e-4} & \multicolumn{1}{c|}{1.5394e-1} & \multicolumn{1}{c|}{1.7698e-1} & \multicolumn{1}{c|}{1.7332e-1} \\ \hline
\end{tabular}

\label{tab: resolution_invariance}
\end{table}

\begin{table}[H]
\centering
\caption{Errors of the generated $H$ for test data with different resolutions of $64 \times 64$, $128 \times 128$, and $256 \times 256$. The training resolution is $64 \times 64$. The first to third rows present the errors when the model is conditioned on bicubically interpolated sparse observations, whereas the fourth to sixth rows present the errors when the model is conditioned on sparsely observed data processed with 2-D convolution.}
\begin{tabular}{|c|cc|cccc|}
\hline
\multirow{2}{*}{\diagbox[dir=SE]{Methods}{Errors}} & \multicolumn{2}{c|}{Test Data}                    & \multicolumn{4}{c|}{Test Errors} \\ \cline{2-7} 
                  & \multicolumn{1}{c|}{$\mathrm{d}x$} & $\mathrm{d}y$ & \multicolumn{1}{c|}{$D_{\text{MSE}}$}            & \multicolumn{1}{c|}{$D_{\text{Fro}}$} & \multicolumn{1}{c|}{$D_{\text{Spe}}$} & \multicolumn{1}{c|}{$D_{\text{Max}}$}   \\ \hline
                  
\multirow{3}{*}{Interpolation} & \multicolumn{1}{c|}{1/64}          & 1/64         & \multicolumn{1}{c|}{1.0850e-4}    & \multicolumn{1}{c|}{3.8475e-2} & \multicolumn{1}{c|}{3.7667e-2} & \multicolumn{1}{c|}{5.6713e-2} \\ \cline{2-7} 
                               & \multicolumn{1}{c|}{1/128}         & 1/128         & \multicolumn{1}{c|}{1.9225e-4} & \multicolumn{1}{c|}{5.4087e-2} & \multicolumn{1}{c|}{4.9166e-2} & \multicolumn{1}{c|}{8.6145e-2} \\ \cline{2-7} 
                               & \multicolumn{1}{c|}{1/256}         & 1/256         & \multicolumn{1}{c|}{2.7885e-4} & \multicolumn{1}{c|}{6.6827e-2} & \multicolumn{1}{c|}{5.9896e-2} & \multicolumn{1}{c|}{1.1095e-1} \\ \hline
\multirow{3}{*}{Convolution}   & \multicolumn{1}{c|}{1/64}          & 1/64         & \multicolumn{1}{c|}{2.3878e-4}    & \multicolumn{1}{c|}{5.5936e-2} & \multicolumn{1}{c|}{5.6179e-2} & \multicolumn{1}{c|}{7.9217e-2}  \\ \cline{2-7} 
                               & \multicolumn{1}{c|}{1/128}         & 1/128         & \multicolumn{1}{c|}{2.8405e-4} & \multicolumn{1}{c|}{7.0255e-2} & \multicolumn{1}{c|}{6.9845e-2} & \multicolumn{1}{c|}{9.5172e-2} \\ \cline{2-7} 
                               & \multicolumn{1}{c|}{1/256}         & 1/256         & \multicolumn{1}{c|}{4.9084e-4} & \multicolumn{1}{c|}{9.4637e-2} & \multicolumn{1}{c|}{9.8297e-2} & \multicolumn{1}{c|}{1.2985e-1} \\ \hline
\end{tabular}

\label{tab: resolution_invariance_H}
\end{table}

\begin{figure}[H]
    \centering
    \includegraphics[width = \linewidth]{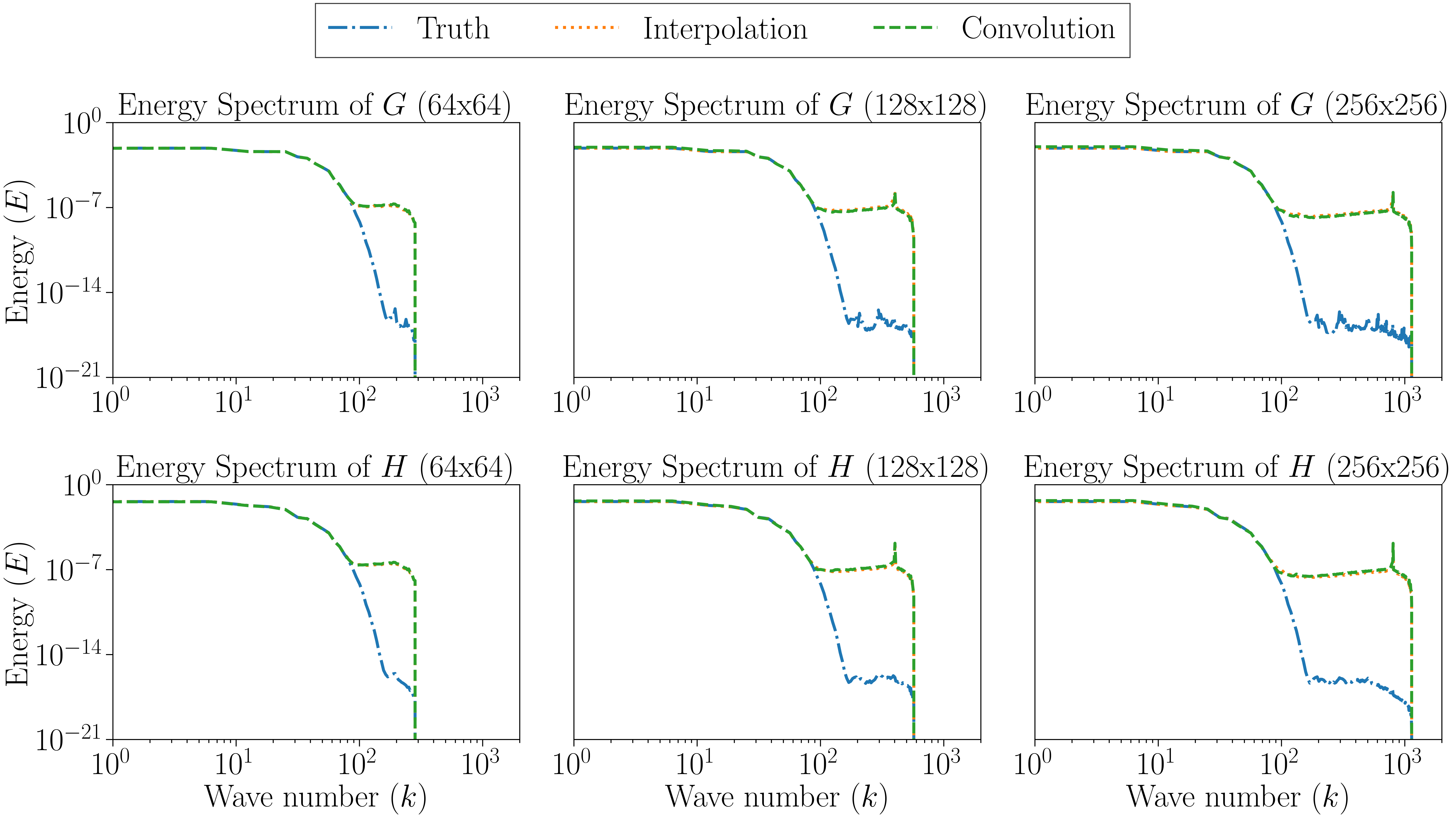}
    \caption{First row: energy spectrum of the generated $G$ with different test resolutions ($64 \times 64$, $128 \times 128$, and $256 \times 256$). Second row: energy spectrum of the generated $H$ with different test resolutions ($64 \times 64$, $128 \times 128$, and $256 \times 256$).}
    \label{fig: TEK_Compare}
\end{figure}

Figure~\ref{fig: TEK_Compare} demonstrates that the generated $G(\mathrm{x}, t)$ and $H(\mathrm{x}, t)$ fields, whether conditioned on interpolated or convoluted sparse information, exhibit a good agreement with the ground truth $G^\dag(\mathrm{x}, t)$ and $H^\dag(\mathrm{x}, t)$ in terms of energy spectra defined in Eq.~\eqref{eqn: energy_spectrum}. This indicates that the model performs well and that both techniques for incorporating sparse information are effective. Additionally, the energy spectra calculated from $G^\dag(\mathrm{x}, t)$ and $H^\dag(\mathrm{x}, t)$ at different resolutions also align closely with one another, suggesting that the proposed conditional score-based model successfully achieves resolution invariance. The main reason for the differences between the true system and the modeled ones in the range of high wave numbers is that the training resolution is only $64 \times 64$, and it is expected that the test results are consistent with the training resolution with lower wave numbers and would not be able to capture the energy spectrum of the true system at higher wave numbers, even with a higher test resolution.

\subsection{Numerical Simulations of the Vorticity using the Trained Models} \label{sec: surrogate}

The trained score-based models can be used as a surrogate closure model for characterizing the difference between the true system in Eq.~\eqref{eqn:true_system} and the modeled system in Eq.~\eqref{eqn:model_system}. In this section, we consider the stochastic 2-D Navier-Stokes in Eq.~\eqref{2dNS} as an example. To simulate the vorticity fields $\omega(\mathrm{x}, t)$, we use the same pseudo-spectral and Crank-Nicolson methods for the data generation process described in \ref{sec: data_generation}. For surrogate modeling, we present two sets of results corresponding to (i) a linear and local operator and (ii) a nonlinear and non-local operator, as detailed in Section~\ref{sec: diffusion_generation} and Section~\ref{sec: nonlinear_generation}, respectively. The first model is trained to conditionally generate the stochastic viscous diffusion term, expressed as $G^\dag(\mathrm{x}, t) = \nu\nabla^2\omega^\dag(\mathrm{x}, t) + 2\beta\xi$, while the second model targets the stochastic convection term $H(\mathrm{x}, t) = -u(\mathrm{x}, t) \cdot \nabla\omega(\mathrm{x}, t) + \beta\xi$. For both simulations, we assume some known information:
\begin{itemize}
        \item Initial vorticity of the true system: $\omega^\dag(\mathrm{x}, t_{0})$.
        \item Deterministic forcing: $ f(\mathrm{x}) = 0.1 (\sin(2\pi (x+y)) + \cos(2\pi (x+y)))$. 
\end{itemize}

\begin{figure}[H]
    \centering
    \includegraphics[width = \linewidth]{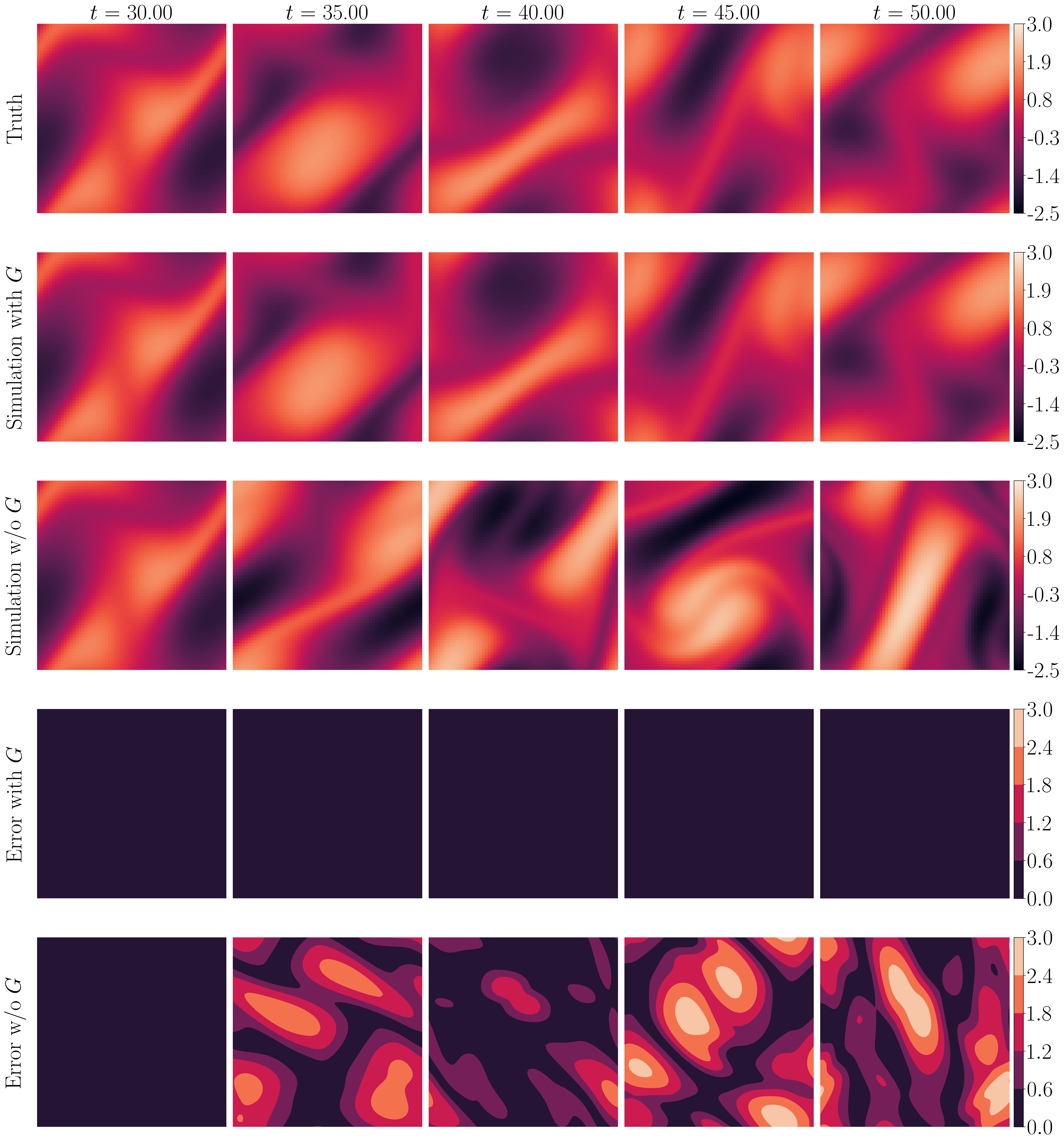}
    \caption{Numerical simulation results of vorticity $\omega$ from 30s to 50s using trained conditional score-based model. Sparse information is upscaled via 2-D convolution. First row: the ground truth vorticity $\omega^\dag$ at various times. Second row: corresponding simulated $\omega$ with generated $G$. Third row: corresponding simulated $\omega$ with $G^\dag$ neglected. Fourth row: absolute error fields between the ground truth of vorticity $\omega^\dag$ and the simulated $\omega$ with $G$. Fifth row: absolute error fields between $\omega^\dag$ and the simulated $\omega$ without $G^\dag$.}
    \label{fig:surrogate_G}
\end{figure}

\begin{figure}[H]
    \centering
    \includegraphics[width = \linewidth]{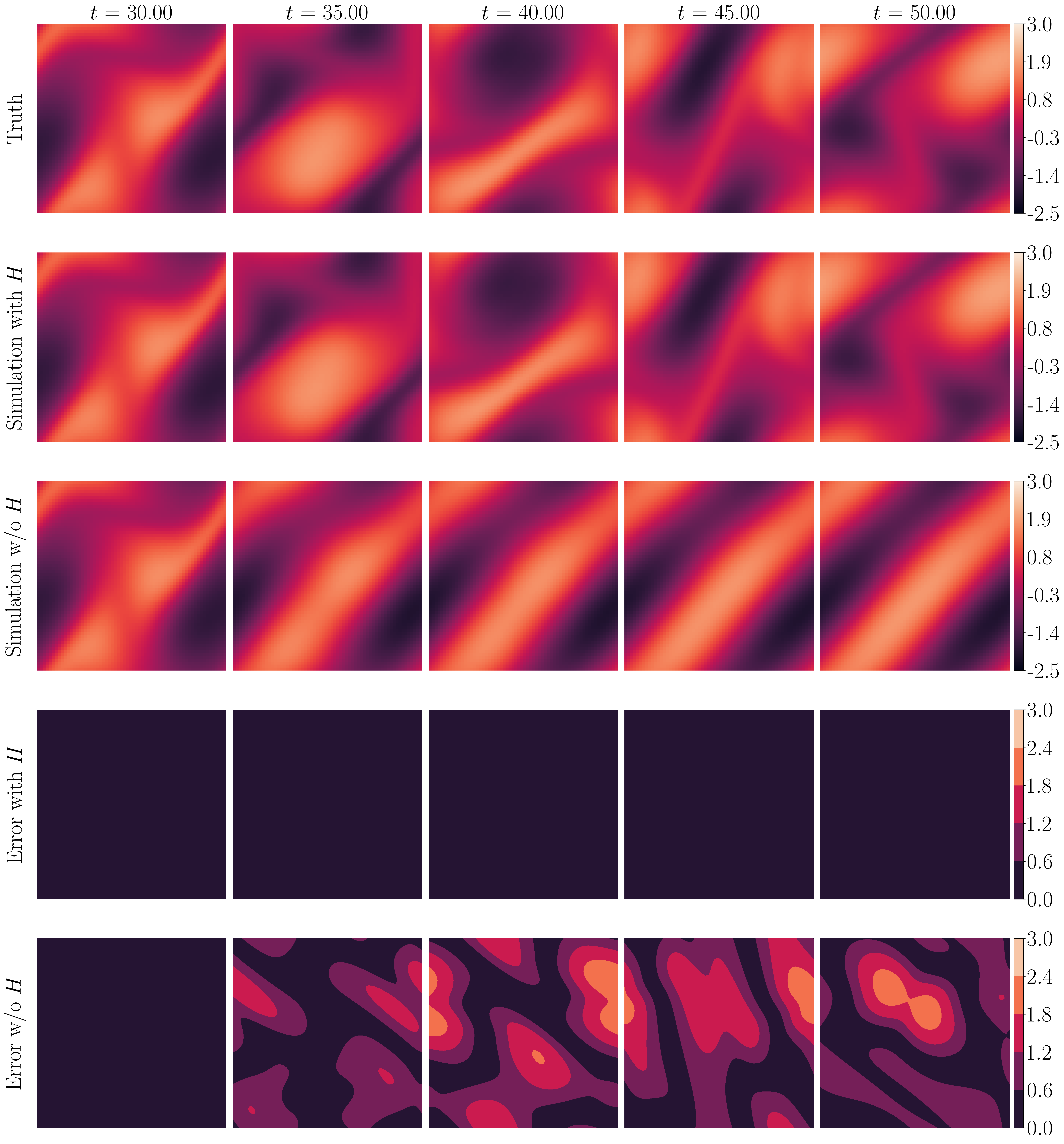}
    \caption{Numerical simulation results of vorticity $\omega$ from 30s to 50s using trained conditional score-based model. Sparse information is upscaled via 2-D convolution. First row: the ground truth vorticity $\omega^\dag$ at various times. Second row: corresponding simulated $\omega$ with generated $H$. Third row: corresponding simulated $\omega$ with $H^\dag$ neglected. Fourth row: absolute error fields between the ground truth of vorticity $\omega^\dag$ and the simulated $\omega$ with $H$. Fifth row: absolute error fields between $\omega^\dag$ and the simulated $\omega$ without $H^\dag$.}
    \label{fig:surrogate_H}
\end{figure}

In general, for a closure term $U(\mathrm{x}, t)$ (representing either $G(\mathrm{x}, t)$ or $H(\mathrm{x}, t)$), the FNO-based score function $s_\theta(\tau, U_\tau, \omega, U^\dag_\text{sparse})$ is trained based on the details described in the previous section, which approximates true score function $\nabla_U \log p\left(U(\mathrm{x}, t)\right)$ in the sampling process of $U(\mathrm{x}, t)$ via the proposed score-based conditional diffusion model. We start with $t_0 = 30$ and aim to simulate the vorticity $\omega(\mathrm{x}, t)$ for $t \in [30, 50]$ with a time step of $\Delta t = 10^{-3}$. Figures~\ref{fig:surrogate_G} and~\ref{fig:surrogate_H} present simulation results under two scenarios: (1) when the ground truth closure terms ($G^\dag$ and $H^\dag$) are neglected and (2) when these terms are generated by the score-based conditional diffusion models and incorporated to close the simulation. The simulations with either missing diffusion or convection exhibit significant deviations from the true system. In contrast, when the closure terms ($G$ or $H$) generated by the conditional diffusion model are included, the simulations align much more closely with the true system, demonstrating the effectiveness of the proposed approach.

It is worth noting that the computational cost can be infeasible for performing full reverse SDE sampling at every step of the Crank-Nicolson update in the numerical simulation of the modeled system. To address this issue and improve efficiency, we first speed up the generation process by starting from a prior distribution that is closer to the ground truth data distribution (i.e., a smaller $\Tau$ for the reverse SDE) and using an adaptive scheme for step sizes, as both discussed in Section~\ref{ssec:method_fast_sampling}. Additionally, improving efficiency while maintaining acceptable error levels can be achieved through sub-sampling the numerical scheme. Specifically, we can perform one instance of reverse SDE sampling every $n$ steps of the Crank-Nicolson update while $n$ is a tunable parameter. For the simulation results with $G$ term in Fig.~\ref{fig:surrogate_G}, we initiate the reverse SDE at $\Tau=0.1$, use 10 adaptive time steps based on the scheduling function in Eq.~\eqref{eqn: adaptive_schedule}, and perform reverse SDE sampling every 5 steps of the Crank-Nicolson update. For $G(\mathrm{x}, t)$ in the intervening steps of the numerical scheme where we opt not to perform reverse SDE sampling, we use an approximation of $G(\mathrm{x}, t_n+\Delta t) = G(\mathrm{x}, t_n) + \beta z$, where $z \sim \mathcal{N}(0, I)$ and $\beta = 5\times 10^{-5}$ is the same noise scale coefficient used in the original 2-D Navier-Stokes equation. In Fig.~\ref{fig:surrogate_H}, the results of the surrogate modeling with $H(\mathrm{x}, t)$ are performed with the same accelerating techniques for generation and simulation.

\begin{figure}[H]
    \centering
    \includegraphics[width = \linewidth]{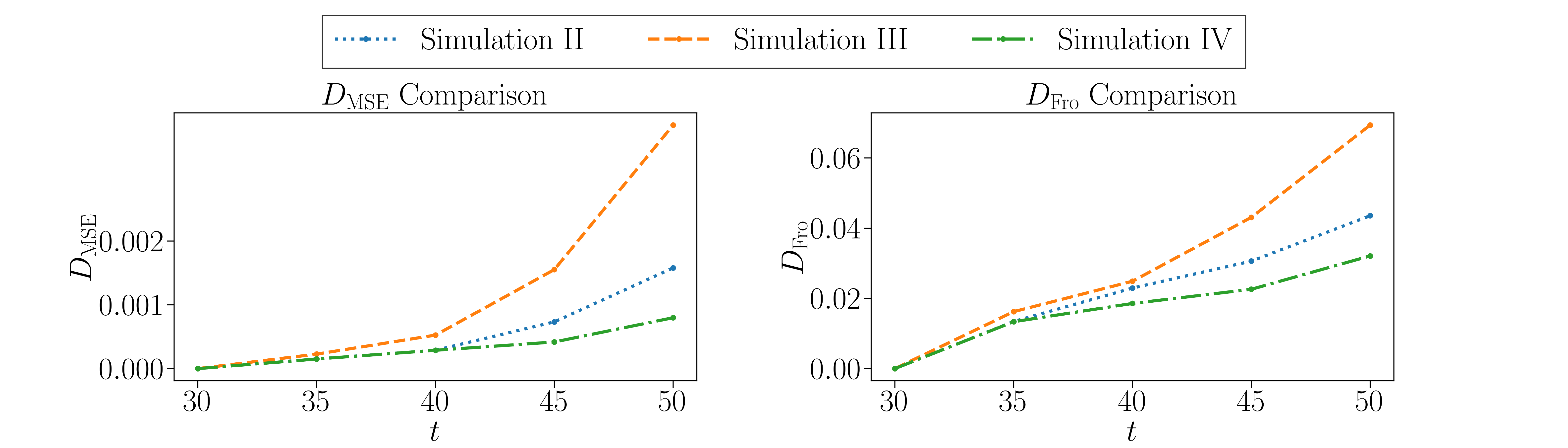}
    \caption{A comparison for the accuracy of simulated vorticity $\omega$ from 30s to 50s with different simulation settings.}
    \label{fig: surrogate_efficiency_G}
\end{figure}

Table~\ref{tab: surrogate_efficiency_G} summarizes the performances and computational costs across four different simulation settings. Simulation I corresponds to the simulated $\omega$ when $G^\dag$ is completely removed, which leads to much larger errors than the other three simulation settings. Simulation II corresponds to the one that utilizes all the speedup techniques introduced above. Simulation III is designed to have similar computational costs as Simulation II. It starts each reverse SDE solving at $\Tau = 0.1$ but utilizes 10 fixed-size time steps, producing less accurate generation of $G$ and subsequently leading to larger errors of $\omega$ than Simulation II, which has adaptive time steps for the reverse SDE. Simulation IV, serving as the baseline method, shows the best accuracy but incurs the highest computational cost due to performing reverse SDE generation at every simulation step, with denoising initiated at $\Tau=1$ and redundant 1000 fixed-size time steps for each sampling process. Figure~\ref{fig: surrogate_efficiency_G} illustrates the error comparison of the numerical simulations that deploy a score-based generative closure term with different sampling strategies, which corresponds to Simulation II, III, and IV. With the costs in Table~\ref{tab: surrogate_efficiency_G} and the errors in Fig.~\ref{fig: surrogate_efficiency_G}, we can see that Simulation II provides a good balance between accuracy and simulation cost, confirming the effectiveness of the proposed fast sampling strategy for deploying diffusion models as stochastic data-driven closure terms in numerical simulations.

\begin{table}[H]
\centering
\caption{A comparison of computational cost, accuracy of generated $G$ terms, and accuracy of simulated vorticity $\omega$ from 30s - 50s. The index I corresponds to the simulation of $\omega$ with $G^\dag$ term neglected. II corresponds to the simulation with a generated $G$ term every 5 physical time steps, and for each generated sample, the denoising process starts at $\mathrm{T}=0.1$ using 10 time steps with adaptive sizes. III corresponds to the simulation with a generated $G$ term every 5 physical time steps, and for each generation, the denoising process starts at $\mathrm{T}=0.1$ using 10 time steps with fixed sizes. IV corresponds to the generated $G$ term at each physical time step, and for each generation, the denoising process starts at $\tau=1$ using 1000 time steps with a fixed size.}

\resizebox{\textwidth}{!}{
\begin{tabular}{|c|c|c|c|c|c|c|c|c|}
\hline
           Simulations        & \begin{tabular}[c]{@{}c@{}}Costs\end{tabular} & Errors & Generated $G$ & $\omega(t = 30)$ & $\omega(t = 35)$ & $\omega(t = 40)$ & $\omega(t = 45)$ & $\omega(t = 50)$ \\ \hline
\multirow{2}{*}{$G_\mathrm{I}$}   & \multirow{2}{*}{7.76s}                                               & $D_{\text{MSE}}$    & NA            &0 &3.9363e-2 &1.6108e-1 &3.8245e-1 &6.6759e-1       \\ \cline{3-9} 
                              &                                                                   & $D_{\text{Fro}}$   & NA            &0 &1.8562e-1 &3.8417e-1 &5.1685e-1 &6.5925e-1        \\ \hline
\multirow{2}{*}{$G_\mathrm{II}$}  & \multirow{2}{*}{276.52s}                                         & $D_{\text{MSE}}$    & 1.4778e-4    &0 &1.5259e-4 &2.8789e-4 &7.3273e-4 &1.5795e-3       \\ \cline{3-9} 
                              &                                                                   & $D_{\text{Fro}}$   & 8.8991e-2          &0 &1.3415e-2 &2.2945e-2 &3.0611e-2 &4.3552e-2          \\ \hline

\multirow{2}{*}{$G_\mathrm{III}$}  & \multirow{2}{*}{274.34s}                                         & $D_{\text{MSE}}$    & 1.7845e-4   &0 &2.2781e-4 &5.2549e-4 &1.5526e-3 &3.8183e-3       \\ \cline{3-9} 
                              &                                                                   & $D_{\text{Fro}}$   & 1.0121e-1       &0 &1.6223e-2 &2.4895e-2 &4.3086e-2 &6.9365e-2           \\ \hline
                              
\multirow{2}{*}{$G_\mathrm{IV}$}   & \multirow{2}{*}{94242.64s}                                       & $D_{\text{MSE}}$    & 6.4787e-5     & 0       &1.5088e-4         & 2.8789e-4           & 4.1774e-4       & 
7.9800e-4       \\ \cline{3-9} 
                              &                                                                   & $D_{\text{Fro 
     }}$   & 6.9606e-2        & 0        &1.3356e-2        & 1.8562e-2                   & 2.2601e-2           & 3.2098e-2           \\ \hline
\end{tabular}}

\label{tab: surrogate_efficiency_G}
\end{table}
\normalsize

\begin{figure}[H]
    \centering
    \includegraphics[width=\linewidth]{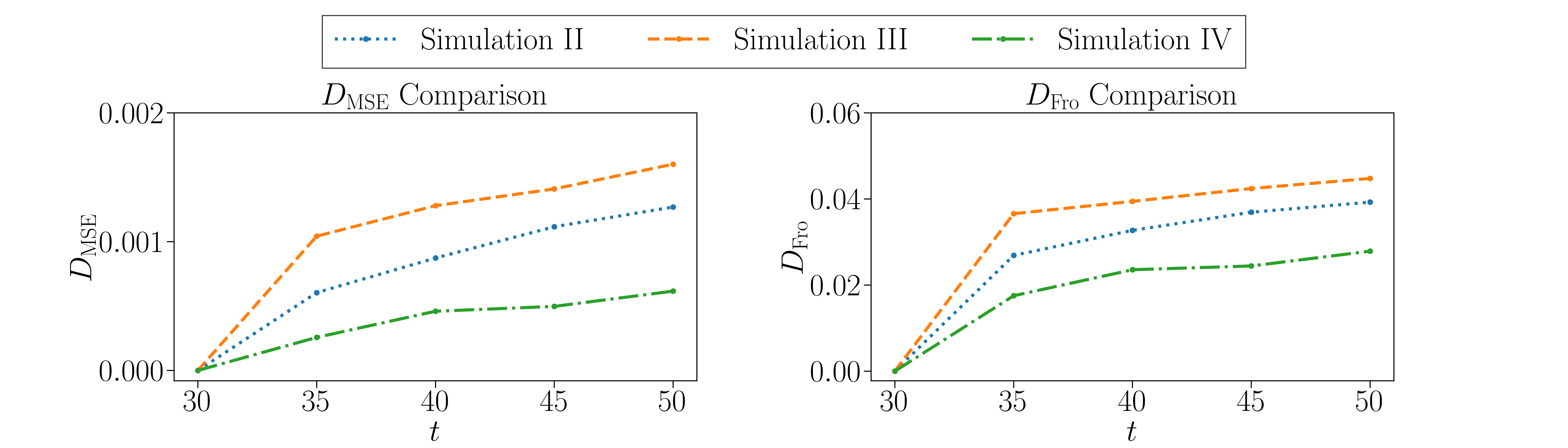}
    \caption{A comparison for the accuracy of simulated vorticity $\omega$ from 30s to 50s with different simulation settings.}
    \label{fig: surrogate_efficiency_H}
\end{figure}

For completeness, we also present the simulation performance in Fig.~\ref{fig:surrogate_H} using the trained score-based model as the closure for the missing stochastic convection terms, \(H(\mathrm{x}, t)\). As summarized in Table~\ref{tab: surrogate_efficiency_H} and Fig.~\ref{fig: surrogate_efficiency_H}, Simulation II consistently achieves the optimal balance between accuracy and computational cost when employing various sampling techniques.

\begin{table}[H]
\centering
\caption{A comparison of computational cost, accuracy of generated $H$ terms, and accuracy of simulated vorticity $\omega$ from 30s - 50s. The index I corresponds to the simulation of $\omega$ with $H^\dag$ term neglected. II corresponds to the simulation with a generated $H$ term every 5 physical time steps, and for each generated sample, the denoising process starts at $\mathrm{T}=0.1$ using 10 time steps with adaptive sizes. III corresponds to the simulation with a generated $H$ term every 5 physical time steps, and for each generation, the denoising process starts at $\mathrm{T}=0.1$ using 10 time steps with fixed sizes. IV corresponds to the generated $H$ term at each physical time step, and for each generation, the denoising process starts at $\tau=1$ using 1000 time steps with a fixed size.}

\resizebox{\textwidth}{!}{
\begin{tabular}{|c|c|c|c|c|c|c|c|c|}
\hline
           Simulations        & \begin{tabular}[c]{@{}c@{}}Costs\end{tabular} & Errors & Generated $H$ & $\omega(t = 30)$ & $\omega(t = 35)$ & $\omega(t = 40)$ & $\omega(t = 45)$ & $\omega(t = 50)$ \\ \hline
\multirow{2}{*}{$H_\mathrm{I}$}   & \multirow{2}{*}{5.81s}                                               & $D_{\text{MSE}}$    & NA            & 0                & 5.2602e-1       & 1.0718     & 1.0098        & 7.3543e-1       \\ \cline{3-9} 
                              &                                                                   & $D_{\text{Fro}}$   & NA            & 0                &7.1536e-1           & 9.5131e-1           & 8.7851e-1         & 7.2650e-1         \\ \hline
\multirow{2}{*}{$H_\mathrm{II}$}  & \multirow{2}{*}{204.73s}                                         & $D_{\text{MSE}}$    & 2.3878e-4    & 0                & 6.0433e-4       & 8.7310e-4       & 1.1163e-3       & 1.2691e-3       \\ \cline{3-9} 
                              &                                                                   & $D_{\text{Fro}}$   & 5.5936e-2        & 0                & 2.6890e-2           & 3.2715e-2           & 3.6923e-2           & 3.9265e-2           \\ \hline

\multirow{2}{*}{$H_\mathrm{III}$}  & \multirow{2}{*}{202.78s}                                         & $D_{\text{MSE}}$    & 2.7745e-4    & 0                &1.0426e-3       &1.2796e-3        & 1.4099e-3       & 1.6015e-3    \\ \cline{3-9} 
                              &                                                                   & $D_{\text{Fro}}$   & 6.6420e-2       & 0                & 3.6581e-2           & 3.9451e-2           & 4.2438e-2           & 4.4774e-2           \\ \hline

\multirow{2}{*}{$H_\mathrm{IV}$}  & \multirow{2}{*}{92534.71s}                                         & $D_{\text{MSE}}$    & 1.5416e-4    & 0                & 2.5687e-4       & 4.5996e-4       & 4.9764e-4       & 6.1615e-4       \\ \cline{3-9} 
                              &                                                                   & $D_{\text{Fro}}$   & 4.7520e-2        & 0                & 1.7523e-2           & 2.3565e-2           & 2.4442e-2           & 2.7875e-2           \\ \hline
\end{tabular}}

\label{tab: surrogate_efficiency_H}
\end{table}

\begin{figure}[H]
    \centering
    \includegraphics[width = \linewidth]{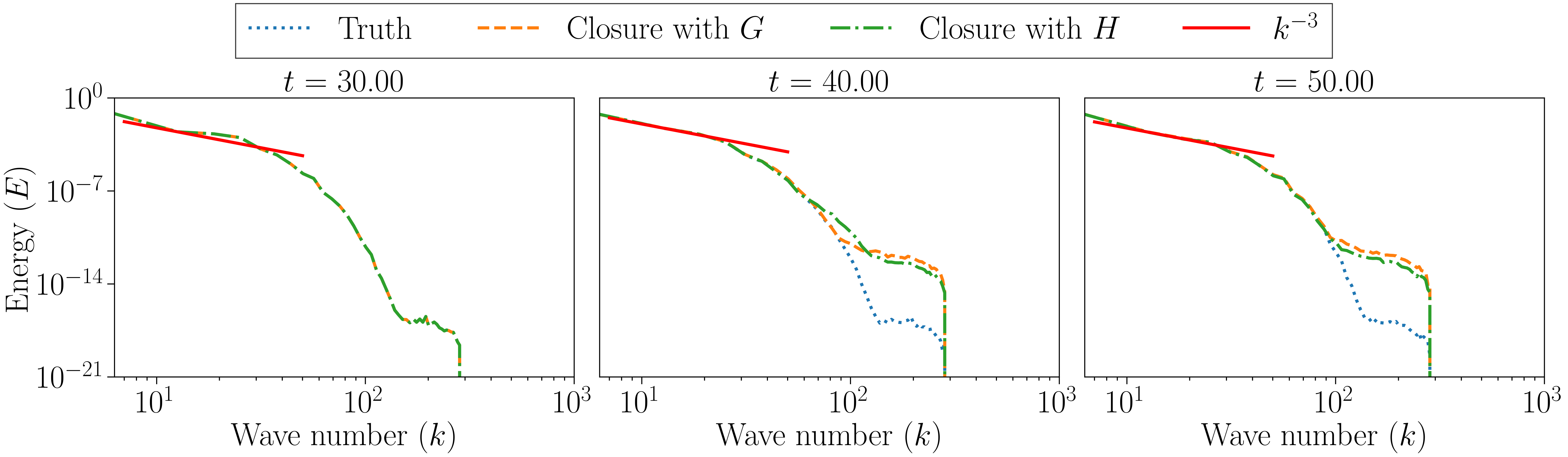}
    \caption{Comparison of energy spectra of 2-D Navier-Stokes equation among the ground truth, reduced-order model with generated closure $G$, and reduced-order model with generated closure $H$ at various times. Only snapshots at times 30, 40, and 50 are shown since the energy spectra exhibit self-similarity during this period, making additional time points redundant.}
    \label{fig:TKE_Closure}
\end{figure}

In Fig.~\ref{fig:TKE_Closure}, we further compare the energy spectrum of the resolved vorticity among the ground truth simulation, the reduced-order model (ROM) with generated closure $G$, and the ROM with generated closure $H$ at various times between 30s and 50s. We include a reference slope $k^{-3}$, which corresponds to the forward enstrophy cascade commonly observed in 2-D turbulence. For the reduced-order runs with closure terms, we follow our second sampling scheme mentioned above as this strategy achieves the best balance between simulation accuracy and computational cost. We see that the simulations with the trained closure models show excellent agreement with the ground truth solution at small wavenumbers, indicating that large-scale flow features are well-corrected after the incorporation of the generated closure terms. The energy spectra at higher wavenumbers exhibit some deviations, which is expected since the training resolution is only $64 \times 64$ and is aligned with the spectral analysis we show in Fig.~\ref{fig: TEK_Compare}. Nonetheless, the overall spectral shape remains consistent with the $k^{-3}$ reference slope, suggesting that the models with trained closures effectively preserve the forward enstrophy cascade characteristic of 2-D turbulent flows.

\normalsize

\section{Conclusions}
Many complex dynamical systems are characterized by a wide range of scales, such as turbulence or the Earth system, for which numerically resolving all the scales is still infeasible in the foreseeable future. Closure models are thus needed to characterize the impact of unresolved scales on the resolved ones. In this work, we present a data-driven modeling framework to build stochastic and non-local closure models based on the conditional diffusion model and neural operator. More specifically, the Fourier neural operator is used to approximate the score function for a score-based generative diffusion model, which captures the conditional probability distribution of the unknown closure term given some dependent information, e.g., the numerically resolved scales of the true system, sparse experimental measurements of the true closure term, and estimation of the closure term from existing physics-based models. Fast sampling algorithms are also investigated to ensure the efficiency of the proposed framework. A comprehensive study is performed on the 2-D Navier-Stokes equation, for which the stochastic viscous diffusion term is assumed to be unknown. Results show that (i) conditioning on dependent information can help achieve a good performance of the diffusion-model-based closure with a limited amount of training data and (ii) employing a neural operator to approximate the score function can help achieve the resolution invariance of the trained model. The proposed framework provides a systematic data-driven approach to building and calibrating stochastic and non-local closure models for some challenging science and engineering applications that lack a scale separation between resolved and unresolved scales, for which deterministic and local closure models are not sophisticated enough to provide a generalizable model that characterizes the contribution of unresolved scales to the resolved ones.

\section*{Data Availability}
The data and trained models that support the findings of this study are available from the corresponding author upon reasonable request. The codes and examples that support the findings of this study are available in the link: \url{https://github.com/AIMS-Madison/SBM_FNO_Closure}.
\section*{Acknowledgments}
X.D., C.C., and J.W. are supported by the University of Wisconsin-Madison, Office of the Vice Chancellor for Research and Graduate Education with funding from the Wisconsin Alumni Research Foundation. X.D. and J.W. are also funded by the Office of Naval Research N00014-24-1-2391.

\bibliographystyle{model1-num-names}
\bibliography{references.bib}

\clearpage

\appendix

\section{Objective Functions of Score Matching}\label{sec:score_matching_proof}
\subsection{Equivalency between ESM and DSM}\label{ssec: proof_ESMDSM}
In this appendix, we first prove that explicit score matching (ESM) and denoising score matching (DSM) have equivalent objectives. Let us consider explicit score matching, which has an objective function
\begin{equation}\label{eqn: ESM}
    \begin{aligned}
        J_\text{ESM}(\theta) &= \mathbb{E}_{U_\tau \sim p(U_\tau)} \left \| \nabla_{U_\tau} \log p(U_\tau) - s_\theta(\tau, U_\tau)\right \|^2_2 \\
        &= \mathbb{E}_{U_\tau \sim p(U_\tau)} \|s_\theta(\tau, U_\tau) \|^2_2 - 2H(\theta) + C_1
    \end{aligned}
\end{equation}
where $C_1 = \mathbb{E}_{U_\tau \sim p(U_\tau)} \| \nabla_{U_\tau} \log p(U_\tau) \|^2_2$ is a constant that does not depend on model parameters $\theta$.

For $H(\theta)$, we have
\begin{equation}
    \begin{aligned}
        H(\theta) &= \mathbb{E}_{U_\tau \sim p(U_\tau)} \left [\left < \nabla_{U_\tau} \log p(U_\tau), s_\theta(\tau, U_\tau) \right > \right]\\
        &= \int_{U_\tau}p(U_\tau) \left < \nabla_{U_\tau} \log p(U_\tau), s_\theta(\tau, U_\tau) \right > \mathrm{d}U_\tau  \\
        &= \int_{U_\tau}p(U_\tau) \left < \frac{\nabla_{U_\tau}p(U_\tau)}{p(U_\tau)}, s_\theta(\tau, U_\tau) \right > \mathrm{d}U_\tau \\
        &= \int_{U_\tau} \left < \nabla_{U_\tau}p(U_\tau), s_\theta(\tau, U_\tau) \right > \mathrm{d}U_\tau \\
        &= \int_{U_\tau} \left < \nabla_{U_\tau}\int_{U_0}p(U_0)p(U_\tau\mid U_0)\mathrm{d}U_0, s_\theta(\tau, U_\tau) \right > \mathrm{d}U_\tau\\
        &= \int_{U_\tau} \left < \int_{U_0}p(U_0)\nabla_{U_\tau}p(U_\tau\mid U_0)\mathrm{d}U_0, s_\theta(\tau, U_\tau) \right > \mathrm{d}U_\tau\\
        &= \int_{U_\tau} \left < \int_{U_0}p(U_0)p(U_\tau\mid U_0)\nabla_{U_\tau}\log p(U_\tau\mid U_0)\mathrm{d}U_0, s_\theta(\tau, U_\tau) \right > \mathrm{d}U_\tau\\
        &= \int_{U_\tau}\int_{U_0} p(U_0) p(U_\tau\mid U_0) \left < \nabla_{U_\tau}\log p(U_\tau\mid U_0), s_\theta(\tau, U_\tau) \right > \mathrm{d}U_0 \mathrm{d}U_\tau \\
        &= \int_{U_\tau}\int_{U_0} p(U_\tau, U_0) \left < \nabla_{U_\tau}\log p(U_\tau\mid U_0), s_\theta(\tau, U_\tau) \right > \mathrm{d}U_0 \mathrm{d}U_\tau \\
        &= \mathbb{E}_{U_\tau \sim p(U_\tau \mid U_0)}\mathbb{E}_{U_0 \sim p(U_0)} \left[ \left < \nabla_{U_\tau}\log p(U_\tau\mid U_0), s_\theta(\tau, U_\tau) \right > \right].
    \end{aligned}
\end{equation}

Substituting this expression for $H(\theta)$ in Eq.~\eqref{eqn: ESM} yields
\begin{equation}
    \begin{aligned}
        J_\text{ESM}(\theta) =& \mathbb{E}_{U_\tau \sim p(U_\tau)} \|s_\theta(\tau, U_\tau) \|^2_2 \\
        &- 2 \mathbb{E}_{U_\tau \sim p(U_\tau \mid U_0)}\mathbb{E}_{U_0 \sim p(U_0)} \left[ \left < \nabla_{U_\tau}\log p(U_\tau\mid U_0), s_\theta(\tau, U_\tau) \right > \right] + C_1.
    \end{aligned}
\end{equation}

For the denoising score matching objective mentioned in Eq.~\eqref{eqn: ESMDSM}, we have
\begin{equation}\label{eqn: DSM}
    \begin{aligned}
        J_\text{DSM}(\theta) =& \mathbb{E}_{U_\tau \sim p(U_\tau \mid U_0)}\mathbb{E}_{U_0 \sim p(U_0)} \left \| \nabla_{U_\tau} \log p(U_\tau \mid U_0) - s_\theta(\tau, U_\tau)\right \|^2_2 \\
        =& \mathbb{E}_{U_\tau \sim p(U_\tau \mid U_0)}\mathbb{E}_{U_0 \sim p(U_0)} \|s_\theta(\tau, U_\tau) \|^2_2 \\
        &-2\mathbb{E}_{U_\tau \sim p(U_\tau \mid U_0)}\mathbb{E}_{U_0 \sim p(U_0)} \left[ \left < \nabla_{U_\tau}\log p(U_\tau\mid U_0), s_\theta(\tau, U_\tau) \right > \right] + C_2,
    \end{aligned}
\end{equation}
where $C_2 = \mathbb{E}_{U_\tau \sim p(U_\tau \mid U_0)}\mathbb{E}_{U_0 \sim p(U_0)} \| \nabla_{U_\tau} \log p(U_\tau \mid U_0) \|^2_2$ is a constant independent of $\theta$.

Viewing Eq.~\eqref{eqn: ESM} and Eq.~\eqref{eqn: DSM}, we notice that
\begin{align}
    J_\text{ESM}(\theta) = J_\text{DSM}(\theta) + C_1 - C_2.
\end{align}
Thus, the two objective functions are equivalent, up to a constant.

\subsection{Conditional Score Matching}\label{ssec: CSM}
For modeling the conditional distribution $p(U \mid \mathrm{y})$, we first recall that the conditional score matching objectives build upon the fact that the forward process of the diffusion model is a Markov chain, meaning that
\begin{align}
    p(U_\tau \mid U_0, \mathrm{y}) = p(U_\tau \mid U_0), 
\end{align}
where $\mathrm{y}$ denotes the conditions. In this paper, $\mathrm{y}$ includes resolved system state $V$, sparse measurements of true $U$, and other dependent information.

Then, for the conditional score matching (CSM), we simply need to further marginalize the conditions $\mathrm{y}$ and get
\begin{equation}
    \begin{aligned}
            J_\text{CSM}(\theta, \mathrm{y}) &= \mathbb{E}_{U_\tau \sim p(U_\tau \mid U_0)}\mathbb{E}_{U_0 \sim p(U_0 \mid \mathrm{y})}\mathbb{E}_{\mathrm{y} \sim p(\mathrm{y})} \left \| \nabla_{U_\tau} \log p(U_\tau \mid U_0, \mathrm{y}) - s_\theta(\tau, U_\tau, \mathrm{y})\right \|^2_2 \\
            &= \mathbb{E}_{U_\tau \sim p(U_\tau \mid U_0)}\mathbb{E}_{\left( U_0,\mathrm{y} \right)\sim p(U_0, \mathrm{y})} \left \| \nabla_{U_\tau} \log p(U_\tau \mid U_0) - s_\theta(\tau, U_\tau,\mathrm{y})\right \|^2_2
    \end{aligned}
\end{equation}

\section{Details of the Numerical Solver}\label{sec: data_generation}
The data for the 2-D Navier-Stokes equation in Eq.~\eqref{2dNS} is generated using the pseudo-spectral method combined with the Crank-Nicolson scheme. 

\subsection{Pseudo-Spectral Solver}\label{ssec: pseudospectral}
Start off, we have initial conditions $\omega(\mathrm{x}, t_0) \sim \mathcal{N}(0, 7^{3/2}(-\Delta + 49I)^{-5/2})$ with periodic boundary conditions, where $\Delta$ is a Laplace operator and $I$ is the identity operator. Operating in Fourier space, we first compute the Fourier transform of the vorticity field at time \(t_0\):
\begin{equation}
    \hat{\omega}(k, t_0) = \mathcal{F}\bigl[\omega(\mathrm{x}, t_0)\bigr],
\end{equation}
where \(\mathcal{F}\) denotes the Fourier transform, \(\hat{\omega}\) denotes the Fourier coefficients of the vorticity field \(\omega\), and 
\begin{equation}
k = (k_x, k_y) = \left(\frac{2\pi n_x}{L}, k_y = \frac{2\pi n_y}{L}\right)
\end{equation}
are the wavenumbers computed from the discrete Fourier indices with $n_x, n_y$ being the integer indices corresponding to the Fourier modes and $L$ the physical length of the domain.

In Fourier space, the Laplacian operator $nabla^2$ is diagonal and acts as
\begin{equation}
    \mathcal{F}\left[\nabla^2 \omega(\mathrm{x}, t)\right] = -\left(k_x^2 + k_y^2\right)\hat{\omega}(k, t).
\end{equation}

To obtain the stream function $\psi$, we solve the Poisson equation in Fourier space:
\begin{equation}
    \nabla^2 \psi = -\omega \quad \Longrightarrow \quad -\left(k_x^2 + k_y^2\right) \hat{\psi}(k) = -\hat{\omega}(k),
\end{equation}
which yields
\begin{equation}
    \hat{\psi}(k) = \frac{\hat{\omega}(k)}{k_x^2 + k_y^2}.
\end{equation}

Finally, the velocity field is obtained from the stream function via the relation
\begin{equation}\label{eqn: fourier_gradient}
    \hat{u}(k) = \left(i k_y\, \hat{\psi}(k),\; - i k_x\, \hat{\psi}(k)\right).
\end{equation}
In this formulation, the factors of \(2\pi\) are already absorbed in \(k_x\) and \(k_y\), ensuring that all operators are expressed in terms of physical wavenumbers.

The gradient of vorticity $\nabla \hat{\omega}$ is also calculated in the Fourier space in a similar fashion as shown in Eq.~\eqref{eqn: fourier_gradient}. Then, 
$\nabla \hat{\omega}$ and $\hat{u}$ are converted back to the physical space via the inverse Fourier transform to calculate the nonlinear convection term, which we denote as
\begin{equation}
    F(\mathrm{x}, t) = u(\mathrm{x}, t) \cdot \nabla \omega(\mathrm{x}, t)
\end{equation}

\subsection{Crank-Nicolson Method}\label{ssec: cranknicolson}
The vorticity field is updated at each time step using the Crank-Nicolson scheme, which is implicit in time and second-order accurate:
\begin{equation}\label{eqn: CrankNicolson1}
    \begin{aligned}
        \hat{\omega}(k, t_{n+1}) &= \frac{\hat{\omega}(k, t_{n}) - \Delta t \hat{F}(k, t_n) + \Delta t \hat{f}(k) + \frac{\Delta t}{2}\nu C \hat{\omega}(k, t_{n}) + \Delta t \beta \hat{\xi}_n}{1+\frac{\Delta t}{2} \nu C } \\
    \end{aligned}
\end{equation}
where $\nu = 10^{-3}$ is the viscosity, $\Delta t = 10^{-3}$ is the time step, $C$ is the Laplacian in Fourier space, $\hat{F}(k, t_n)$ is the Fourier transform of the nonlinear convection term, and $\hat{f}$ is the Fourier transform of the deterministic forcing term.

Recall that in Section~\ref{sec: numerical_results}, we have the ground truth of the correction term $G^\dag(\mathrm{x}, t) = \nu\nabla^2\omega(\mathrm{x}, t) + 2\beta\xi$. The above equation can be rewritten as
\begin{equation}\label{eqn: CrankNicolson2}
    \begin{aligned}
        \hat{\omega}(k, t_{n+1}) &= \frac{\hat{\omega}(k, t_{n}) - \Delta t \hat{F}_n + \Delta t \hat{f}(k) + \frac{\Delta t}{2} \hat{G^\dag}(k, t_n)}{1+\frac{\Delta t}{2} \nu C}\\
        &\approx \frac{\hat{\omega}(k, t_{n}) - \Delta t \hat{F}_n + \Delta t \hat{f}(k) + \frac{\Delta t}{2} \hat{G}(k, t_n)}{1+\frac{\Delta t}{2} \nu C},
    \end{aligned}
\end{equation}
which represents using our surrogate model $G$ to simulate the 2-D Navier-Stokes system shown in Section~\ref{sec: surrogate}.

\section{Model Architecture and Training Details}\label{sec: train_detail}
The conditional score-based diffusion model $s_\theta(\tau, U_\tau, V, U^\dag_{\text{sparse}})$ employs multiple parallel FNO pipelines, each dedicated to processing a specific input variable. The first FNO pipeline processes the noisy closure terms $U_\tau$ along with the forward SDE time parameter $\tau$, where $\tau$ is encoded via Gaussian random features as in Eq.~\eqref{eqn: GRF} and transformed through a dense layer to a high-dimensional space, rendering the diffusion model SDE time-dependent. Two additional FNO pipelines handle the conditional inputs: the current system state $V$ and the sparse information $U^\dag_{\text{sparse}}$, which is originally defined on a $16 \times 16$ grid; during training, $U^\dag_{\text{sparse}}$ is upsampled to $64 \times 64$ using bicubic interpolation or a 2-D convolution with a uniform averaging filter, and during testing, it is upscaled to $128 \times 128$ or $256 \times 256$ to demonstrate resolution invariance. In each FNO pipeline, the spatial input is augmented with normalized grid coordinates and processed through four consecutive Fourier layers, each comprising a physical-space convolution with a $1 \times 1$ kernel and a spectral convolution that transforms the input to the Fourier domain via FFT, applies learnable complex weights to first 16 frequency modes, and reverts to the spatial domain using an inverse FFT. These layers efficiently capture long-range spatial dependencies. The outputs of these convolutions are summed and activated with the GELU function before passing into the next Fourier layer. Finally, the feature maps from all pipelines are concatenated along the channel dimension and refined through a transformation network, consisting of additional $1 \times 1$ convolutions with GELU activations, to produce the final score field.

In our experiments, 8,000 instances were used for training and 2,000 instances for testing. Each instance comprises three components: the target closure term \( U \), the corresponding resolved system state \( V \), and the upscaled sparse information \( U^\dagger_{\text{sparse}} \), all provided at a resolution of \( 64 \times 64 \). The model was trained for 1,000 epochs using the Adam optimizer with an initial learning rate of 0.001, which was reduced by a factor of 2 every 200 epochs. A minibatch size of 80 was employed during training. The model consists of 697,237 parameters and all computations were executed on a single NVIDIA RTX 4090 GPU (24GB). The entire training process required approximately 80 minutes.

\section{Training and Sampling Algorithms}\label{sec: algorithms}
For illustrating the algorithms to train the score-based generative model and to sample from it, we consider the VE SDE shown in Eq.~\eqref{eqn: VESDE} as a concrete example. Its corresponding Gaussian perturbing kernel and reverse SDE are given in Eq.~\eqref{eqn: transitionkernelparam} and Eq.~\eqref{eqn: conditional_reverseSDE}, and the training and sampling algorithms are presented below.

\begin{minipage}{0.45\textwidth}
    \begin{algorithm}[H]
        \caption{Training}
        \begin{algorithmic}[1]
        \REPEAT
            \STATE $U_0, \mathrm{y} \sim p(U_0, \mathrm{y})$
            \STATE $\tau \sim \mathcal{U}(0, \Tau)$
            \STATE $\epsilon \sim \mathcal{N}(0, I)$
            \STATE $U_\tau = U_0 + \sqrt{\frac{1}{2 \ln{\sigma}}\left(\sigma^{2\tau} - 1\right)} \epsilon$
            \STATE
            $\mathcal{L}_\tau(\theta) =$\\ $D_\text{MSE}\left(s_\theta(\tau, U_\tau, \mathrm{y}), \sqrt{\frac{2}{\sigma^{2\tau} - 1} \ln{\sigma}} \epsilon \right)$
            \STATE
            $\theta' = \theta - \nabla_\theta \mathcal{L}_\tau(\theta)$
        \UNTIL{statistical convergence}
        \end{algorithmic}
    \end{algorithm}
\end{minipage}
\hfill
\begin{minipage}{0.45\textwidth}
    \begin{algorithm}[H]
        \caption{Sampling}
        \begin{algorithmic}[1]
        \STATE $U_\Tau \sim \mathcal{N}\left(0, \frac{1}{2 \ln{\sigma}}\left(\sigma^{2\Tau} - 1\right) I\right)$
        \STATE Set $\tau_{\max} = \Tau$, $\tau_{\min} = 10^{-3}$, $\rho = 7$
        \FOR{$i = 0$ to $N-1$}
            \STATE $\tau_i = \left( \tau_{\max}^{\frac{1}{\rho}} + \frac{i}{N-1} \left( \tau_{\min}^{\frac{1}{\rho}} - \tau_{\max}^{\frac{1}{\rho}} \right) \right)^{\rho}$
        \ENDFOR
        \STATE Set $\tau_N = 0$
        \FOR{$i = 0$ to $N-1$}
            \STATE $\Delta \tau = \tau_{i} - \tau_{i+1}$
            \STATE $\epsilon \sim \mathcal{N}(0, I)$
            \STATE $U_{\tau_{i+1}} = U_{\tau_i} + \sigma^{2\tau_i}s_\theta(\tau_i,  U_{\tau_i}, \mathrm{y})\Delta \tau + \sigma^{\tau_i} \sqrt{\Delta \tau}\epsilon$
        \ENDFOR
        \end{algorithmic}
    \end{algorithm}
\end{minipage}

% \clearpage
\end{document}